\documentclass[letterpaper]{article} 
\usepackage{aaai2027}  
\usepackage[hyphens]{url}  
\usepackage{graphicx} 
\usepackage{placeins}
\usepackage{natbib}  
\usepackage{caption} 
\usepackage{multirow}
\usepackage{array}
\usepackage{makecell}
\usepackage{pifont}
\usepackage{xcolor}
\usepackage{amsmath,amssymb}
\newcommand{\yesmark}{\ding{51}}

\usepackage{algorithm}
\usepackage{algorithmic}

\usepackage{newfloat}
\usepackage{listings}
\DeclareCaptionStyle{ruled}{labelfont=normalfont,labelsep=colon,strut=off} 
\floatstyle{ruled}
\newfloat{listing}{tb}{lst}{}
\floatname{listing}{Listing}

\usepackage{booktabs}

\title{WorldSimProbe: Diagnosing Simulator Faithfulness in Action-Conditioned World Models for Embodied Manipulation}
\author{
Peterson Co\textsuperscript{\rm 1,\rm 2,\rm 3}\textsuperscript{*}\textsuperscript{\textdagger},
Sicheng Hu\textsuperscript{\rm 1,\rm 2}\textsuperscript{*},
Chunxuan Jiao\textsuperscript{\rm 1,\rm 2}\textsuperscript{*},
Hongyang Cheng\textsuperscript{\rm 2},
Yulin Luo\textsuperscript{\rm 1},\\
Yijie Xu\textsuperscript{\rm 2,\rm 4},
Sixiang Chen\textsuperscript{\rm 1},
Zhongxia Zhao\textsuperscript{\rm 2},
Zihao Wang\textsuperscript{\rm 5},
DaFeng Chi\textsuperscript{\rm 3},\\
Peidong Liu\textsuperscript{\rm 3},
YuTong Chen\textsuperscript{\rm 2,\rm 6},
Henghua Liu\textsuperscript{\rm 2,\rm 6},
Zhihao Yuan\textsuperscript{\rm 3},
Huizhu Jia\textsuperscript{\rm 1},\\
Yuzheng Zhuang\textsuperscript{\rm 3},
Tianle Zhang\textsuperscript{\rm 3},
Liang Lin\textsuperscript{\rm 3},
Huajie Tan\textsuperscript{\rm 2}\textsuperscript{\textdagger},
Shanghang Zhang\textsuperscript{\rm 1}\textsuperscript{\ding{41}}
}
\affiliations{
\textsuperscript{\rm 1}State Key Laboratory of Multimedia Information Processing, School of Computer Science, Peking University,
\textsuperscript{\rm 2}Evophys.ai,
\textsuperscript{\rm 3}Joy Future Academy, JD\\
\textsuperscript{\rm 4}The University of Sydney,
\textsuperscript{\rm 5}The Hong Kong University of Science and Technology,
\textsuperscript{\rm 6}Beijing Institute of Technology\\
\textsuperscript{\ding{41}} Corresponding author: \texttt{shanghang@pku.edu.cn}
}

\affiliations{}

\newcommand{\taskheading}[1]{\par\medskip\noindent\textbf{#1}\par\nobreak\smallskip}

\nocopyright
\begin{document}

\maketitle

\begingroup

\renewcommand{\thefootnote}{}
\footnotetext{\footnotesize
\textsuperscript{*} Equal contribution.
\textsuperscript{\textdagger} Project leaders.
\textsuperscript{\ding{41}} Corresponding author.
\textsuperscript{\rm 1} State Key Laboratory of Multimedia Information Processing, School of Computer Science, Peking University
\textsuperscript{\rm 2} EvoPhys AI
\textsuperscript{\rm 3} Joy Future Academy, JD
\textsuperscript{\rm 4} The University of Sydney
\textsuperscript{\rm 5} The Hong Kong University of Science and Technology
\textsuperscript{\rm 6} Beijing Institute of Technology.
Correspondence to: Shanghang Zhang \texttt{<shanghang@pku.edu.cn>}.
}
\addtocounter{footnote}{-1}
\endgroup

\begin{abstract}
Action-conditioned world models (ACWMs) promise to provide embodied AI with scalable predictive simulators for planning, policy evaluation, and data generation. Realizing this promise requires precise action-conditioned transitions rather than merely plausible outputs. Yet their applicability remains difficult to establish because prevailing evaluations emphasize visual quality, task outcomes, or coarse rollout-level responsiveness without directly testing simulator fidelity. To address this gap, we evaluate ACWMs through the observable capabilities expected of physical simulators. Accordingly, we formalize \textbf{Observable Simulator Contract}, a minimal contract that any action-conditioned physical simulator should satisfy: supplied actions must induce corresponding agent motion, and environment responses must be grounded in that realized motion. To operationalize this contract, we introduce \textbf{WorldSimProbe}, comprising five controlled suites spanning local control sensitivity, global trajectory variation, source-diverse actions, interaction grounding, and dynamics. Suite-specific evaluators assess simulator-relative calibration, dense action-to-motion correspondence, false-interaction grounding, and primitive-level dynamics. We evaluate six open-source ACWMs on more than \(18{,}000\) instances across RoboTwin, ManiSkill, and LIBERO. WorldSimProbe reveals systematic action-realization degradation across control variation, structured failures in interaction grounding and dynamics, and benchmark signals consistent with human judgments and downstream outcomes. Together, this capability-based framework provides a transparent, and standardized paradigm for diagnosing ACWM simulator fidelity beyond coarse, task-directed evaluation. Code and data available here: \url{https://evophys.com/WorldSimProbe/}
\end{abstract}
\begin{figure}[!ht]
    \centering
    \includegraphics[width=\columnwidth]{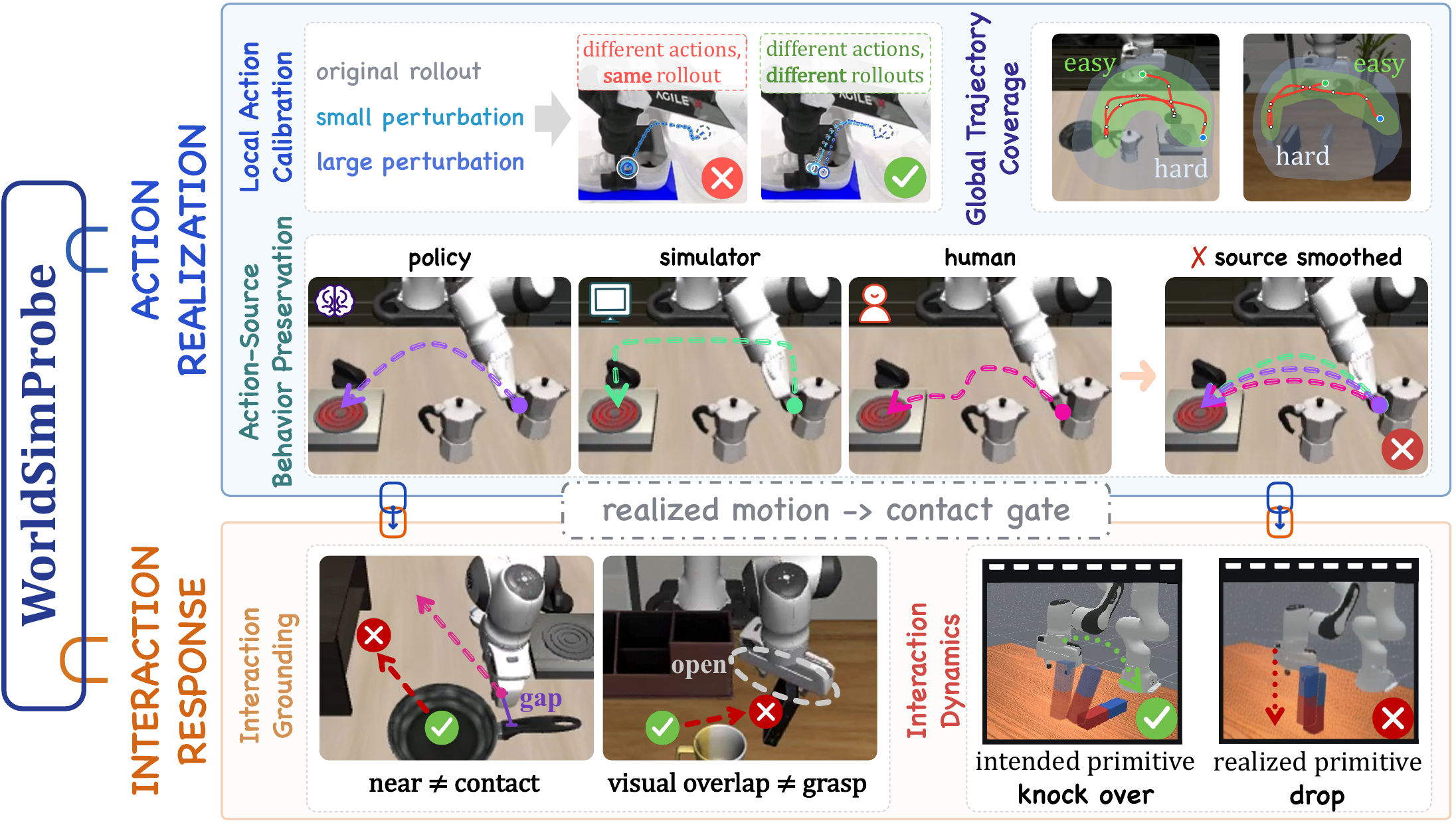}
    \caption{Overview of the simulator-faithfulness chain from supplied action to agent motion and environment response.}
    \label{fig:intro_overview}
\end{figure}

World models increasingly support embodied intelligence and Physical AI by predicting future scene evolution~\cite{yang2024unisim,bruce2024genie,agarwal2025cosmos}. Action-conditioned world models (ACWMs) extend this capability to robotic manipulation by predicting visual futures under supplied action streams~\cite{zhu2025irasim,zhu2025uwm,ctrlworld2025,dreamdojo2026,oscar2026}. Their defining requirement is therefore not merely to generate plausible futures, but to generate the particular future physically induced by the supplied action in the current scene.

Recent benchmarks have extended ACWM evaluation beyond visual fidelity toward action reliability, physical plausibility, and downstream utility~\cite{shang2026worldarena,jiang2026robowmbench,yang2026mirabench,mmbench22026}. However, existing evaluations often remain limited in diagnostic capability: action-following is typically assessed at the rollout level or within task-specific control distributions, providing insufficient coverage of physically valid action variations. Moreover, they rarely examine whether predicted environment responses are causally supported by the motion induced by the supplied actions. As a result, a rollout may appear action-aware or physically plausible while lacking faithful action execution or physically grounded interactions.

Although ACWMs support diverse downstream applications, their reliability ultimately depends on a common foundation: faithful simulation of action-conditioned transitions. We therefore introduce \textit{\textbf{Observable Simulator Contract}}, the minimal requirement for physical simulators: \textit{valid actions should produce corresponding agent motions, and environmental responses should emerge from physically grounded interactions.} As illustrated in Figure~\ref{fig:intro_overview}, this defines a simulator chain from action realization to interaction dynamics, including realized agent motion and environment response.


Within this simulator chain, action realization itself is inherently multi-dimensional rather than monolithic. It requires sensitivity to fine-grained perturbations, diversity in trajectory realizations, and robustness across heterogeneous control distributions and execution styles. Beyond generating plausible motion, a faithful simulator must ensure that realized motions induce interactions consistent with scene context, contact constraints, and mechanism dynamics throughout the entire process. Therefore, reaching a correct final state through unsupported interactions or incorrect dynamics is insufficient. Reliable simulation requires preserving three fundamental properties: \textit{action-to-motion correspondence, interaction grounding, and interaction-dynamics fidelity}. 
Performance in a single regime does not guarantee robustness across others, motivating evaluations that systematically probe the full simulator chain rather than isolated rollout success.

To this end, we introduce \textit{\textbf{WorldSimProbe}}, a diagnostic benchmark that evaluates ACWM faithfulness through controlled interventions along the simulator chain. WorldSimProbe operationalizes the \textit{\textbf{Observable Simulator Contract}} through five evaluation suites: Local Action Calibration, Global Trajectory Coverage, Action-Source Behavior Preservation, Interaction Grounding, and Interaction Dynamics. The first three characterize complementary aspects of action realization, while the latter two evaluate whether environment responses are physically supported by realized motions. Each suite combines simulator-executable interventions with targeted evaluators, enabling failure localization beyond a single aggregate rollout score.

Our contributions are summarized as follows:

\begin{itemize}

\item We define the \textit{\textbf{Observable Simulator Contract}}, formalizing simulator faithfulness through observable correspondences between supplied actions, realized motions, interactions, and environment responses.

\item We introduce \textit{\textbf{WorldSimProbe}}, a diagnostic benchmark with five controlled suites along the simulator chain, comprising over 18,000 instances across three simulators and diverse embodiments. Further, we develop multi-dimensional action-fidelity metrics and interaction evaluators to diagnose distinct failure modes, including action calibration, action-motion correspondence, unsupported interactions, and interaction dynamics.

\item We also conduct a systematic study of six mainstream ACWM baselines, revealing failures hidden by existing coarse evaluations, including action compression, degraded generalization beyond task-specific controls, unsupported interactions, and inconsistent dynamics.

\end{itemize}

\section*{2. Related Work}

\begin{table}[!t]
\centering
{\small
\renewcommand{\arraystretch}{0.96}
\renewcommand{\yesmark}{{\color[HTML]{006B3C}\ding{51}}}
\setlength{\tabcolsep}{0.65mm}
\begin{tabular*}{\columnwidth}{@{\extracolsep{\fill}}lccccc@{}}
\toprule
& \multicolumn{3}{c}{\textbf{Action Realization}}
& \multicolumn{2}{c}{\textbf{Interaction}} \\
\cmidrule(lr){2-4}\cmidrule(lr){5-6}
\textbf{Benchmark}
& \shortstack{\textbf{Explicit}\\\textbf{Action}}
& \shortstack{\textbf{Local}\\\textbf{Global}}
& \shortstack{\textbf{Source}\\\textbf{Diverse}}
& \shortstack{\textbf{Causal}\\\textbf{Probe}}
& \shortstack{\textbf{Interaction}\\\textbf{Decomp.}} \\
\midrule
What-If World & -- & -- & -- & \yesmark & -- \\
WorldSimBench & -- & -- & -- & -- & -- \\
MiraBench & \yesmark & -- & -- & \yesmark & -- \\
RoboWM-Bench & -- & -- & -- & -- & -- \\
WMBench & \yesmark & -- & \yesmark & -- & -- \\
WorldArena & \yesmark & -- & \yesmark & -- & -- \\
ACWM-Phys & \yesmark & -- & -- & \yesmark & -- \\
\textbf{WorldSimProbe} & \yesmark & \yesmark & \yesmark & \yesmark & \yesmark \\
\bottomrule
\end{tabular*}
}
\caption{Capability comparison of embodied world-model benchmarks. Local--Global tests graded perturbations and divergent trajectories; Source Diverse uses multiple control sources; Causal Probe applies controlled interventions; Interaction Decomp. separates interaction occurrence from dynamics. Checks denote explicit evaluation.}
\label{tab:benchmark_comparison}
\end{table}

\textbf{Output-oriented and Indirect Evaluation.}
Video-oriented benchmarks assess language- or instruction-conditioned
generations through visual fidelity, semantic alignment, motion correctness,
task completeness, and physical plausibility
~\cite{hu2025ewmbench,li2025worldmodelbench,deng2026rethinking}.
Some benchmarks introduce language-level interventions or infer actions from
generated videos without directly supplying executable controls
~\cite{jiang2026robowmbench,cai2026whatifworld,li2026robotrustbench,qin2025worldsimbench}.
A single language instruction may admit many valid action trajectories, while
inferred actions depend on an auxiliary estimator; both therefore provide
coarse or indirect evidence of action following.

\vspace{0.5em}

\textbf{Explicit Action-conditioned Evaluation.}
Benchmarks that condition on explicit robot actions enable fine-grained
interventions on control magnitude, timing, and trajectory, and direct
measurement of realized motion
~\cite{yang2026mirabench,xue2026acwmphys,mmbench22026}.
However, these evaluations do not jointly trace fidelity across control scales
and distributions from supplied action through realized motion to environment
response.

\vspace{0.5em}

\textbf{Downstream-utility Evaluation.}
Downstream-utility evaluations assess world models through policy evaluation,
policy ranking, online interaction, or synthetic-data generation
~\cite{shang2026worldarena,shang2026worldarena2,tseng2026sc3eval,
quevedo2026worldgym,li2025worldeval,li2026dworldeval,
wang2026interactiveworldsimulator,team2026gigaworld}.
Although they directly assess application value, aggregate success or reward
provides limited failure localization, while the emphasis on task-directed
trajectories can underrepresent counterfactual, recovery, failure-inducing, and
out-of-distribution controls.
Table~\ref{tab:benchmark_comparison} summarizes this progression. WorldSimProbe traces two linked simulator transitions: supplied action to realized agent motion, and realized motion to environment response.

\section*{3. Simulator Faithfulness Framework}

\subsection*{3.1 Observable Simulator Contract}

Given an initial scene observation \(x_t\) and a supplied action stream \(a_{t:t+H}\), an action-conditioned world model generates a future rollout \(\hat{x}_{t:t+H}\). The action stream specifies the intervention whose physical consequences the rollout must realize. We decompose the scene into an initial agent state \(r_t\) and environment state \(e_t\), and the generated rollout into the realized agent motion \(\hat{r}_{t:t+H}\) and environment response \(\hat{e}_{t:t+H}\). Exogenous scene changes are absent or fixed within each controlled evaluation instance, allowing the environment response to be attributed to the supplied action through the realized agent motion.

Simulator faithfulness requires two linked consistency conditions. First, action-realization consistency requires the generated agent motion to correspond to the supplied action under the current scene:
\[
\hat{r}_{t:t+H} \approx \Phi_R(r_t,e_t,a_{t:t+H}).
\]
Second, interaction-response consistency requires the generated environment response to be physically supported by the realized agent motion and initial environment state:
\[
\hat{e}_{t:t+H} \approx \Phi_E(e_t,\hat{r}_{t:t+H}).
\]
Here, \(\Phi_R\) and \(\Phi_E\) denote the ideal but generally unobserved action-realization and interaction-response operators. WorldSimProbe does not require direct access to these operators; instead, each suite constructs controlled interventions that test their observable consequences. A rollout satisfies the simulator contract only when both links hold jointly: the supplied action is realized as corresponding agent motion, and environment changes are mediated by that motion and its contact events. This distinguishes simulator faithfulness from visual plausibility or task consistency alone.

\begin{figure}[H]
  \centering
  \includegraphics[width=\columnwidth]{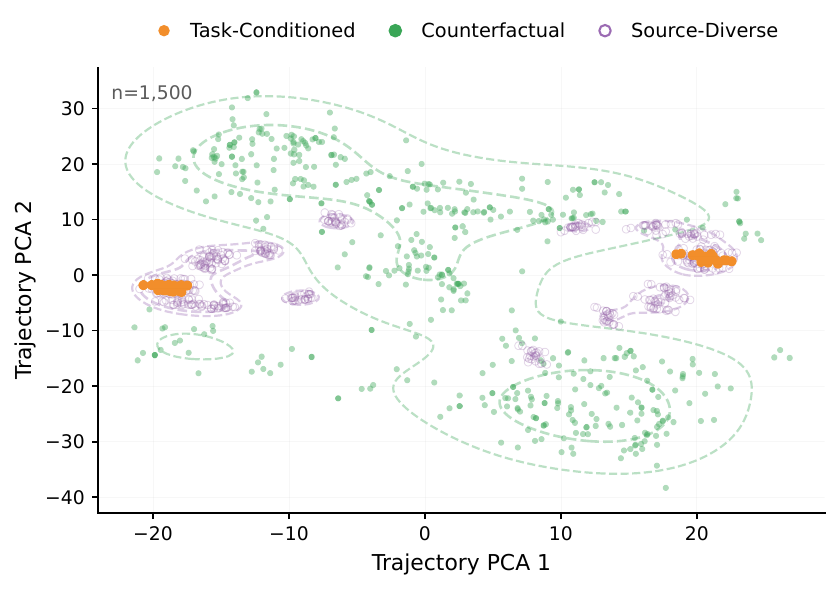}
  \caption{Representative feasible-action coverage for the RoboTwin \texttt{handover\_mic} task ($n=1{,}500$). PCA projects 12-dimensional arm-motion trajectories, excluding gripper dimensions, for task-conditioned, counterfactual, and source-diverse actions.}
  \label{fig:feasible-action-coverage}
\end{figure}

\begin{figure*}[t]
    \centering
    \includegraphics[width=\textwidth]{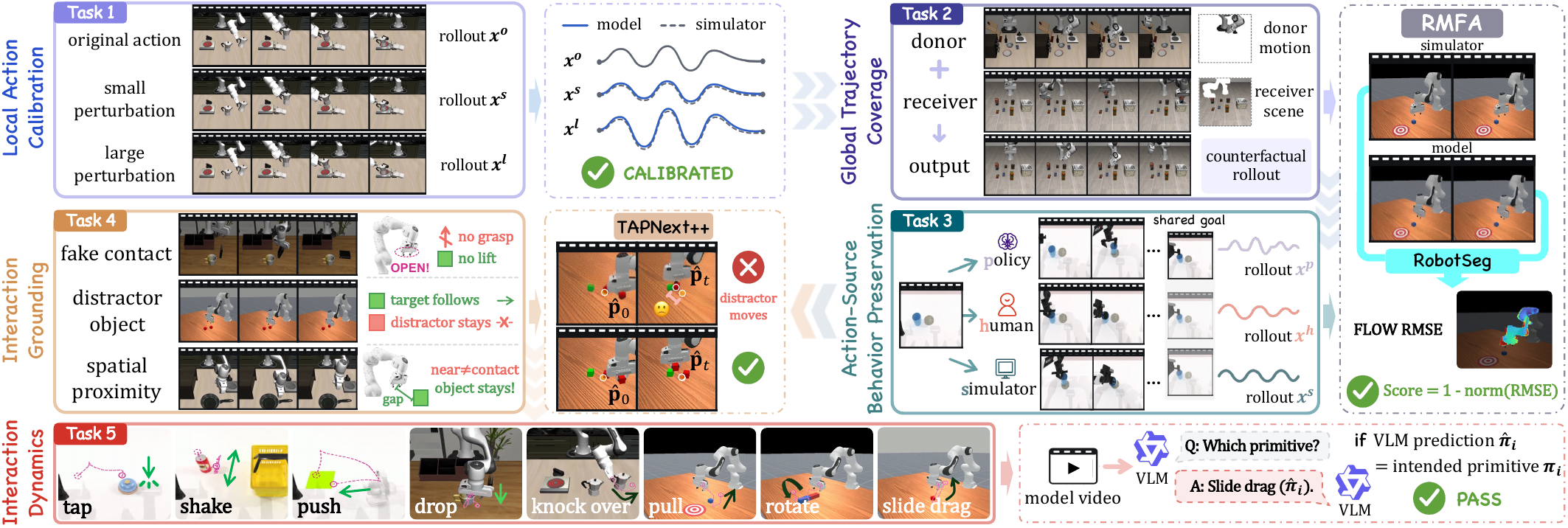}
    \caption{Operational overview of WorldSimProbe. The five suites progress from local action calibration through global and source-specific action realization to interaction grounding and interaction dynamics, using suite-specific evaluators to localize failures along the simulator chain.}
    \label{fig:operational_overview}
\end{figure*}

\subsection*{3.2 Feasible Motion Coverage for Action Realization}

The action-realization link of the simulator contract imposes both a correspondence and a coverage requirement: a simulator must map each supplied action to its physically induced motion, and this mapping must remain valid across all controls executable from the current state. Figure~\ref{fig:feasible-action-coverage} illustrates this expectation gap: physical simulators span the full distribution of feasible trajectories, whereas ACWMs are typically trained and evaluated on narrower task-associated subsets. For an initial state \((r_t,e_t)\), let \(\mathcal{A}_{\mathrm{phys}}(r_t,e_t)\) contain all action streams executable under the scene geometry, embodiment, controller limits, and evaluation horizon, and let \(\mathcal{A}_{\mathrm{task}}\subseteq\mathcal{A}_{\mathrm{phys}}\) denote task-protocol support. Even when \(\mathcal{A}_{\mathrm{task}}\) includes both successful and failed trajectories, such outcome diversity does not establish coverage beyond the task distribution. WorldSimProbe addresses this gap through local perturbations, globally divergent but executable trajectories, and heterogeneous control sources, testing action realization independently of task success.

\section*{4. WorldSimProbe: A Diagnostic Benchmark}

WorldSimProbe instantiates the simulator contract through five diagnostic suites progressing from supplied action to realized agent motion and environment response (Figure~\ref{fig:operational_overview}). The first three broaden action-realization tests from local perturbations to global and source-diverse controls; the final two extend evaluation to interaction grounding and dynamics, enabling stage-specific failure localization.

\paragraph{Task 1: Local Action Calibration.}
In a physical simulator, changing the supplied action should change the resulting rollout according to the physical consequence of that intervention. This correspondence should hold even when the intervention does not alter task semantics or success. Task 1 isolates this local regime by holding the task and outcome fixed while applying fine-grained action perturbations that induce subtle but measurable trajectory changes. A faithful ACWM should reproduce the simulator's calibrated response rather than merely react monotonically to larger perturbations.

Starting from a successful action trajectory, we construct an original stream and two variants by perturbing one non-gripper action dimension over a fixed window at two magnitudes. The simulator episode seed, each model's diffusion seed, and the perturbation dimension, direction, and window are shared across variants. Using temporally aligned full-video MSE \(D(\cdot,\cdot)\), we define a common evaluation set using only simulator references, retaining triplets for which all variants remain task-valid and successful and \(0<D(x_i^0,x_i^s)<D(x_i^0,x_i^\ell)\).

Let \(\hat{x}_i^0,\hat{x}_i^s,\hat{x}_i^\ell\) denote the generated rollouts and \(x_i^0,x_i^s,x_i^\ell\) their simulator references. We define
\[
r_i=
\frac{D(\hat{x}_i^0,\hat{x}_i^s)}
     {D(\hat{x}_i^0,\hat{x}_i^\ell)},
\qquad
r_i^*=
\frac{D(x_i^0,x_i^s)}
     {D(x_i^0,x_i^\ell)}.
\]

Because ACWMs differ in visual quality and baseline pixel error, absolute MSE responses may not be directly comparable across models. The ratios instead capture within-model response scaling, which we compare with the simulator scaling through
\[
S_i=
\operatorname{clip}_{[0,1]}
\begin{cases}
r_i/r_i^*, & r_i \leq r_i^*,\\[2pt]
(1-r_i)/(1-r_i^*), & r_i>r_i^*.
\end{cases}
\]

The score reaches one when the model reproduces the simulator's relative response scaling and decreases as the two relationships diverge; reversed or degenerate scaling receives zero. We report \(100\,\mathbb{E}_i[S_i]\) as the Task 1 score. This oracle-relative score reduces sensitivity to model-specific visual-error scale and evaluates local response calibration; Tasks 2 and 3 separately assess dense action-to-motion correspondence.

\paragraph{Task 2: Global Trajectory Coverage.}
Moving from local perturbations to global trajectory variation, Task 2 evaluates action realization beyond task-associated trajectories through cross-task receiver--donor replay. For receiver scene \(i\) with initial state \((r_i^R,e_i^R)\), we execute an action stream \(a_i^D\) sampled from a different donor task in the receiver simulator, producing the reference \(x_i^{\mathrm{cf},*}\). The model receives the same receiver observation and action stream and generates \(\hat{x}_i^{\mathrm{cf}}\). This construction expands the evaluated action support while retaining a simulator-grounded reference for each intervention.

We evaluate action realization using \emph{Robot-Masked Flow
Alignment} (RMFA). After temporally aligning the generated and
simulator-reference rollouts, we estimate their dense optical flow
using a pretrained estimator~\cite{morimitsu2025dpflow}. A robot-arm
segmentation model~\cite{mei2026robotseg} is applied only to the
simulator reference, defining a common evaluation region independent
of segmentation quality in generated videos. For temporal window \(w\),
we retain reference robot pixels exhibiting active motion:
\[
\mathcal{M}_{i,w}=\bigl\{p\in\mathcal{M}_{\mathrm{arm}}(x_i^{\mathrm{cf},*}):\|F_w(x_i^{\mathrm{cf},*})(p)\|_2>\tau\bigr\}.
\]
where \(F_w(\cdot)\) denotes the optical-flow field over \(w\). Within
this mask, we compute the generated-to-reference flow error and
reference-motion magnitude:
\[
E_{i,w}
=
\operatorname{RMS}_{p\in\mathcal{M}_{i,w}}
\left\|
F_w(\hat{x}_i^{\mathrm{cf}})(p)
-
F_w(x_i^{\mathrm{cf},*})(p)
\right\|_2,
\]
\[
R_{i,w}
=
\operatorname{RMS}_{p\in\mathcal{M}_{i,w}}
\left\|
F_w(x_i^{\mathrm{cf},*})(p)
\right\|_2.
\]
We then define
\[
S_{i,w}
=
100\max\!\left(
0,\,
1-\frac{E_{i,w}}{\max(R_{i,w},c)}
\right),
\]
where \(c\) limits sensitivity to low-motion flow noise. The Task~2
score averages \(S_{i,w}\) over overlapping windows and evaluation
instances. Because RMFA compares generated and reference flow vectors at corresponding reference-arm pixels, it is sensitive to errors in robot-motion direction, magnitude, and spatial alignment.

\paragraph{Task 3: Action-Source Behavior Preservation.}
Unlike the procedural controls in Tasks 1 and 2, Task 3 tests whether action
realization preserves source-specific behavior. We collect successful and
unsuccessful task-directed trajectories from multiple human teleoperators and
early- and late-training policy checkpoints, spanning variation in speed,
smoothness, corrective behavior, and execution strategy. For each trajectory,
the corresponding simulator rollout \(x_i^{\mathrm{src},*}\) provides the
reference; given the same initial observation and action stream, the model
generates \(\hat{x}_i^{\mathrm{src}}\). We compute RMFA over active
reference-arm pixels and average normalized scores across temporal windows and
instances. Higher scores indicate source-specific motion preservation; lower scores indicate reversion to canonical execution.
 Figure~\ref{fig:feasible-action-coverage} illustrates the broader, multimodal coverage induced by counterfactual and source-diverse trajectories in a representative RoboTwin task.

\paragraph{Task 4: Interaction Grounding.}
Having evaluated action realization, we examine the next simulator-chain link: whether realized motion should produce an agent--environment interaction. A simulator determines this from the scene state and executed controls; interaction should occur only when the realized trajectory and control state satisfy the required physical preconditions.

Preliminary qualitative analysis of sampled generations across all six models found missed responses to supported contact to be rare; interaction-grounding failures were instead dominated by \textbf{contact hallucination}, where a model generates agent motion or an environment response as though contact occurred despite being unsupported by the supplied action and scene. Accordingly, we construct controlled no-contact action--scene pairs by adjusting object positions, modifying action streams, or introducing duplicate objects, while retaining visual and spatial cues commonly associated with interaction. Simulator replay verifies that the supplied action, under the constructed scene configuration, does not establish valid contact.

Given the same initial observation and action stream, the model generates a rollout for the validated no-contact case. Episode metadata provides the evaluated object's location, which initializes a TAPNext++ point tracker~\cite{jung2026tapnextpp} at \(\hat{\mathbf{p}}_{i,0}\). Let \(\hat{\mathbf{p}}_{i,t}\) denote its position at frame \(t\). We measure the maximum tracked displacement,
\[
d_i=\max_{t\in\{1,\ldots,T\}}\left\|\hat{\mathbf{p}}_{i,t}-\hat{\mathbf{p}}_{i,0}\right\|_2,
\qquad
S_i=\mathbf{1}\!\left[d_i\leq\tau_{\mathrm{move}}\right].
\]
The Task 4 score is \(S_{\mathrm{T4}}=\frac{100}{N}\sum_{i=1}^{N}S_i\). Because the evaluated object should remain stationary, displacement above \(\tau_{\mathrm{move}}\) indicates that the model has activated an unsupported interaction. This endpoint captures contact hallucination arising either from agent motion that creates spurious contact or from an environment response generated without valid contact.
A motion gate verifies robot-arm centroid displacement across three frames; rollouts that fail the gate receive \(S_i=0\) and remain in the Task 4 denominator.

\begin{table*}[!t]
\centering
{\small
\setlength{\tabcolsep}{2pt}
\renewcommand{\arraystretch}{1.08}
\begin{tabular*}{\textwidth}{
@{\extracolsep{\fill}}l
ccc@{\hspace{5pt}}
ccc@{\hspace{5pt}}
ccc@{\hspace{5pt}}
ccc@{\hspace{5pt}}
ccc@{\hspace{5pt}}
ccc@{}
}
\toprule
\multirow{2}{*}{\textbf{Model}} &
\multicolumn{3}{c}{\textbf{T1 Local}} &
\multicolumn{3}{c}{\textbf{T2 Global}} &
\multicolumn{3}{c}{\textbf{T3 Source}} &
\multicolumn{3}{c}{\textbf{T4 Ground.}} &
\multicolumn{3}{c}{\textbf{T5 Dynamics}} &
\multicolumn{3}{c}{\textbf{Overall}} \\
\cmidrule(lr){2-4}
\cmidrule(lr){5-7}
\cmidrule(lr){8-10}
\cmidrule(lr){11-13}
\cmidrule(lr){14-16}
\cmidrule(l){17-19}
& \textbf{RT} & \textbf{MS} & \textbf{LB}
& \textbf{RT} & \textbf{MS} & \textbf{LB}
& \textbf{RT} & \textbf{MS} & \textbf{LB}
& \textbf{RT} & \textbf{MS} & \textbf{LB}
& \textbf{RT} & \textbf{MS} & \textbf{LB}
& \textbf{RT} & \textbf{MS} & \textbf{LB} \\
\midrule
\multicolumn{19}{@{}l}{\textit{Action-injection models}} \\
\addlinespace[2pt]
IRASim
& 49.4 & 62.3 & 54.0 & 49.4 & 80.0 & 54.7 & 45.5 & 73.6 & 47.9
& 56.0 & 42.2 & 68.2 & 20.5 & 16.5 & 22.4 & 44.2 & 54.9 & 49.4 \\
Ctrl-World
& \textbf{52.4} & 60.0 & 62.7 & 51.1 & 78.0 & \textbf{61.7} & 48.8 & \textbf{78.2} & 50.4
& \textbf{70.0} & 49.9 & \textbf{77.5} & 23.6 & \textbf{20.1} & \textbf{25.0} & 49.2 & 57.2 & \textbf{55.5} \\
BWM
& 38.9 & 44.7 & 48.9 & 45.5 & 77.8 & 58.8 & 35.4 & 75.1 & 43.7
& 51.0 & 43.5 & 63.5 & 20.6 & 19.3 & 23.8 & 38.3 & 52.1 & 47.7 \\
DreamDojo
& 35.2 & 62.9 & 69.5 & 45.7 & 81.4 & 41.5 & 39.2 & 72.6 & 50.3
& 49.0 & 47.2 & 65.3 & 26.9 & 19.9 & 23.5 & 39.2 & 56.8 & 50.0 \\
\addlinespace[3pt]
\multicolumn{19}{@{}l}{\textit{Unified action--video models}} \\
\addlinespace[2pt]
LingBot-VA
& 48.0 & \textbf{68.2} & \textbf{77.0} & \textbf{64.4} & \textbf{82.5} & 60.2 & \textbf{62.0} & 74.7 & \textbf{51.0}
& 53.4 & \textbf{76.3} & 57.8 & \textbf{30.2} & 15.8 & 22.4 & \textbf{51.6} & \textbf{63.5} & 53.7 \\
Cosmos-3-Nano
& 35.2 & 32.6 & 67.3 & 46.2 & 76.9 & 55.9 & 35.3 & 73.4 & 47.4
& 39.6 & 37.1 & 61.1 & 21.7 & 19.6 & \textbf{25.0} & 35.6 & 47.9 & 51.3 \\
\bottomrule
\end{tabular*}
}
\caption{Main WorldSimProbe results across RoboTwin (RT)~\cite{mu2025robotwin}, ManiSkill (MS)~\cite{tao2024maniskill3}, and
LIBERO (LB)~\cite{liu2023libero}. Scores are reported on a \(0\)–\(100\) scale. Overall is the unweighted macro-average across T1–T5 within each simulator and is provided as a summary; individual suite scores remain the primary diagnostic results. Baselines include
IRASim~\cite{zhu2025irasim}, Ctrl-World~\cite{ctrlworld2025},
BWM~\cite{boundlesswm2026}, DreamDojo~\cite{dreamdojo2026},
LingBot-VA~\cite{lingbotva2026}, and
Cosmos-3-Nano~\cite{cosmos32026}. Higher is better; the best result in each
task--simulator column is bold.}
\label{tab:main_results}
\end{table*}

\paragraph{Task 5: Interaction Dynamics.}
If an interaction occurs, a simulator must model its dynamics
under the supplied action. We evaluate this capability through
representative interaction primitives that capture characteristic
relationships between agent motion and environment response
and form the building blocks of successful and unsuccessful
rollouts. These primitives are drawn from simulator tasks,
which provide reproducible controls and references, and from
interactions observed in unsuccessful trajectories, which extend
coverage beyond success-oriented protocols. The set comprises
eight primitives: push, pull, drag, rotate, shake, tap, knock-over,
and drop. Environment response alone cannot distinguish these
primitives; for example, push, pull, and drag can all produce
planar displacement but differ in the agent's motion relative to
the object. For each primitive $\pi \in \Pi$, we use Qwen3-VL-8B
\cite{bai2025qwen3} as the VLM judge and provide a predefined
specification $q_\pi$ describing the expected agent motion and
object response. We validate simulator-defined primitive labels
using three human annotators on a stratified subset and retain
only reference rollouts confirmed by at least two annotators.
On this human-verified set, the VLM judge agrees with the
majority human label on 94--97\% of instances across simulators
and is then applied to every corresponding model generation.

Given the specifications and generated agent–environment rollout, the VLM predicts the realized primitive:
\[
\hat{\pi}_i =
\mathrm{VLM}\!\left(
\hat{x}_{i,t:t+H},
\{q_\pi\}_{\pi\in\Pi}
\right).
\]
The Task 5 score is
\[
S_{\mathrm{T5}} =
\frac{100}{N}\sum_{i=1}^{N}
\mathbf{1}\!\left[\hat{\pi}_i=\pi_i\right],
\]
where \(N\) is the number of retained instances.



\begin{figure}[!t]
    \centering
    \includegraphics[width=\columnwidth]{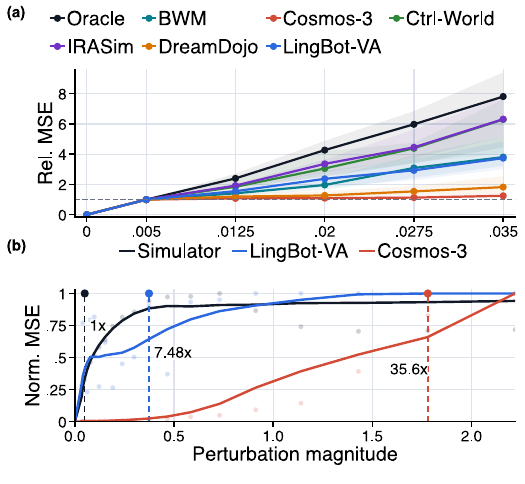}
    \caption{Action calibration and downstream failure onset. (a) Mean relative perturbation--response curves with 95\% confidence intervals across 50 RoboTwin tasks; the horizontal dashed line marks the smallest nonzero perturbation baseline. (b) StackCube case study in ManiSkill; vertical dashed lines mark each simulator or model’s first persistent failure.}
    \label{fig:action_calibration}
\end{figure}

\section*{5. Experiments and Results}

We investigate four questions: \textbf{RQ1}, how simulator faithfulness varies across models, platforms, and stages of the simulator chain; \textbf{RQ2}, how robustly action-to-motion correspondence is preserved across sampled simulator-executable control variations; \textbf{RQ3}, how agent--environment interaction failures vary across grounding conditions and dynamics; and \textbf{RQ4}, whether WorldSimProbe provides empirically valid evaluations with practical downstream relevance.

\subsection*{5.1 Experimental Setup}
We evaluate six representative open-source ACWMs on controlled rollouts from
RoboTwin, ManiSkill, and LIBERO. The final benchmark comprises approximately
\(5{,}500\) RoboTwin, \(6{,}600\) ManiSkill, and \(6{,}500\) LIBERO evaluation
instances, totaling more than \(18{,}000\). They span action-injection architectures with
dedicated action modules and unified action--video architectures that jointly
model actions and observations. Each model is trained separately on each
simulator's official split using its released recipe. We modify only the action dimensionality and input image size to match each
simulator interface and configure the two unified models for action-conditioned,
video-only output; all other training and inference settings retain their
released defaults. Per instance, models share the initial observation, native action trajectory,
and three diffusion seeds; scores average the resulting stochastic generations.
Control frequency, frame rate, and horizon match the simulator reference; evaluator thresholds are fixed from reference data across models.

\subsection*{5.2 Overall Simulator Faithfulness (RQ1)}
Table~\ref{tab:main_results} shows substantial cross-platform ranking
consistency (mean pairwise Spearman \(\rho=0.695\)). LingBot-VA leads
RoboTwin and ManiSkill, whereas Ctrl-World leads LIBERO and ranks second
elsewhere. Lower-ranked models vary more across platforms, while both
architectural families span higher and lower ranks, indicating that architecture
alone does not explain simulator faithfulness. Suite-level rankings nevertheless
differ: Ctrl-World leads Interaction Grounding on RoboTwin and LIBERO and achieves the best cross-platform macro score, while LingBot-VA leads on ManiSkill. Ctrl-World leads overall only on LIBERO. This stage-specific variation shows that strength at one simulator-chain stage does not imply overall fidelity and enables targeted failure localization along the chain.

\subsection*{5.3 Robustness of Action-to-Motion Correspondence (RQ2)}

\noindent\textbf{Local action calibration.}
Figure~\ref{fig:action_calibration}(a) shows that the simulator oracle responds
progressively to increasing perturbations, whereas model responses exhibit
varying degrees of attenuation, indicating that detecting action changes does
not guarantee calibrated response scaling. Panel~\ref{fig:action_calibration}(b)
connects this gap to the \emph{action-failure threshold}, the smallest
perturbation beyond which failure persists. In the StackCube case study,
LingBot-VA crosses at \(0.374\) (\(7.48\times\) the simulator threshold),
substantially earlier than Cosmos-3-Nano at \(1.78\) (\(35.6\times\)), although
both lag the simulator boundary of \(0.05\). This links stronger local
calibration to more faithful failure onset.

\begin{figure}[!t]
  \centering
  \includegraphics[width=\columnwidth]{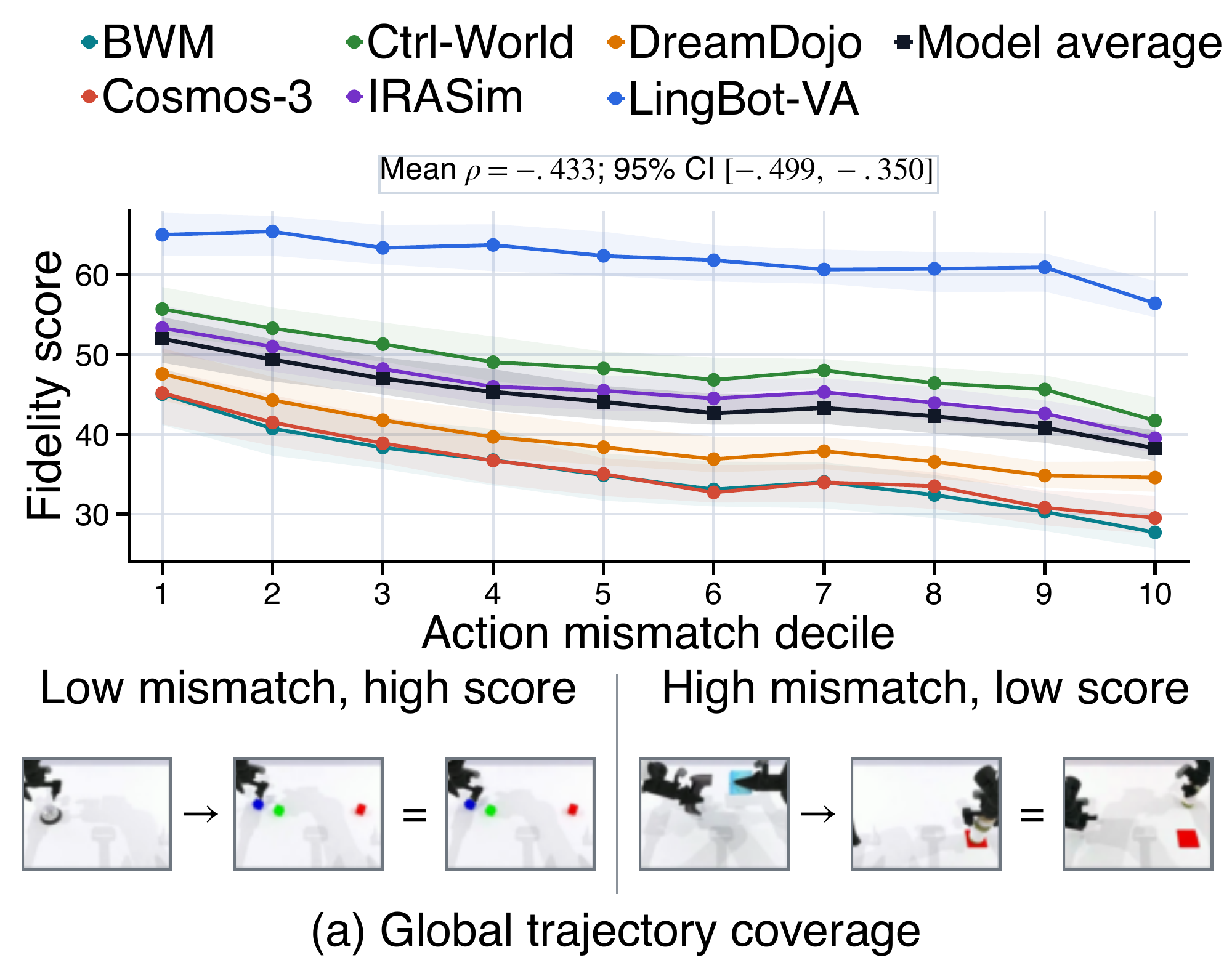}
  \par
    \includegraphics[width=0.96\columnwidth]{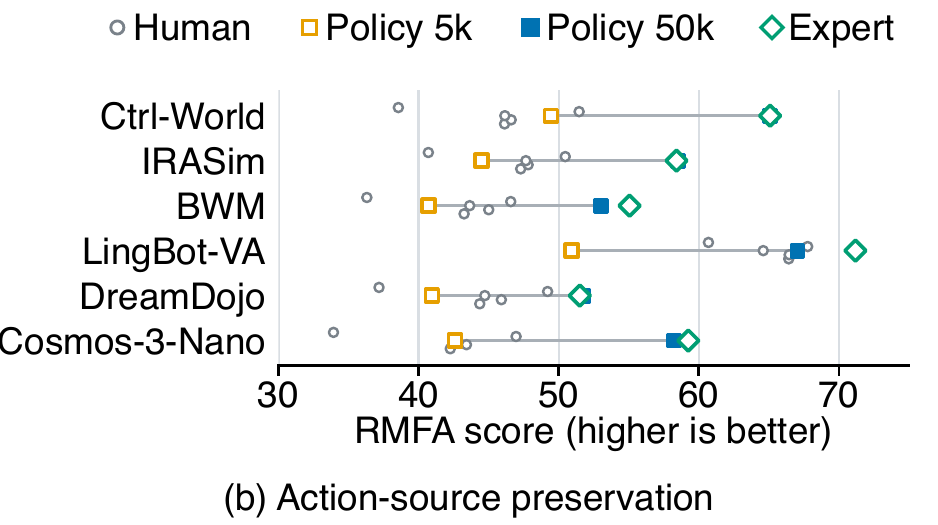}
  \caption{Action realization beyond local perturbations on RoboTwin. \textbf{(a)}
  fidelity across receiver--donor motion-mismatch deciles with representative
  low- and high-mismatch cases. \textbf{(b)} RMFA across human, policy-checkpoint,
  and expert trajectories.}
  \label{fig:global_source_behavior}
\end{figure}

\noindent\textbf{Global trajectory coverage.}
To characterize where global coverage breaks down, we rank receiver–donor pairs by the mismatch between donor robot-arm motion and motion typical of the receiver task, then divide them into ten levels.
Figure~\ref{fig:global_source_behavior} (a) shows that action-realization
fidelity declines overall for all six models as receiver--donor motion mismatch
increases, with an average Spearman correlation of \(\rho=-0.433\). This pattern
is consistent with models relying increasingly on scene- or task-associated
motion priors when supplied controls depart from familiar trajectories.

\noindent\textbf{Action-source behavior preservation.}
Figure~\ref{fig:global_source_behavior} (b) disaggregates Task 3 by control source. All six
models score \(10.8\)--\(16.1\) points higher on late- than early-policy
checkpoints, while human trajectories exhibit consistent operator-dependent
variation. Thus, action realization degrades not only for global counterfactuals
but also for task-directed trajectories with unfamiliar execution
characteristics, limiting applications that require preservation of supplied
behavior.

\begin{table}[!t]
\centering
{\small
\setlength{\tabcolsep}{2.2pt}
\renewcommand{\arraystretch}{0.98}
\begin{tabular*}{\columnwidth}{@{\extracolsep{\fill}}lrrrr@{}}
\toprule
\textbf{Model}
& \shortstack{\textbf{Distr.}\\\textbf{obj.}}
& \shortstack{\textbf{False}\\\textbf{contact}}
& \shortstack{\textbf{Spatial}\\\textbf{prox.}}
& \textbf{Mean} \\
\midrule
IRASim        & 70.3 & 40.3 & 55.9 & 55.5 \\
Ctrl-World    & 79.5 & \textbf{56.1} & 61.8 & \textbf{65.8} \\
BWM           & 72.6 & 41.1 & 44.3 & 52.7 \\
DreamDojo     & 72.9 & 30.9 & 57.7 & 53.8 \\
LingBot-VA    & \textbf{83.3} & 40.3 & \textbf{63.9} & 62.5 \\
Cosmos-3      & 71.5 & 23.0 & 43.4 & 45.9 \\
\bottomrule
\end{tabular*}
}
\caption{False-interaction grounding, macro-averaged across RoboTwin,
ManiSkill, and LIBERO. Mean averages the three triggers.}
\label{tab:false_interaction_grounding}
\end{table}

\begin{figure}[!t]
\centering
\includegraphics[width=\columnwidth]{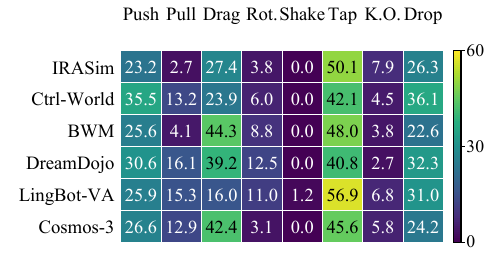}
\caption{Interaction-primitive fidelity, macro-averaged across RoboTwin,
ManiSkill, and LIBERO. Values are percentages; shading encodes magnitude.}
\label{fig:interaction_primitive_fidelity}
\end{figure}

\subsection*{5.4 Structure of Agent--Environment Interaction Failures (RQ3)}

\noindent\textbf{Interaction grounding.}
We evaluate three no-contact settings: spatial proximity places an object near
the trajectory; appearance-induced false contact preserves contact-like visual
evidence but removes contact-enabling control; and distractor cases add plausible
objects that should remain stationary. Across models and platforms, grounding
is strongest for distractors (\(75.0\)) and proximity (\(54.5\)) but drops for
false contact (\(38.6\)) (Table~\ref{tab:false_interaction_grounding}). Models resist distractors and proximity better than visual contact cues lacking control support.

\noindent\textbf{Interaction dynamics.}
Fidelity varies more across primitives than models
(Figure~\ref{fig:interaction_primitive_fidelity}). Tap is strongest
(\(40.8\)--\(56.9\)), followed by drag, drop, and push; pull, rotate, and
knock-over are weaker, while shake is nearly absent (\(0.0\)--\(1.2\)). The
narrow model-average range (\(17.7\)--\(21.8\)) indicates shared
primitive-specific biases. Human validation on model generations yields 93–95\% agreement, supporting these findings. Spanning successful and failed rollouts, these primitives expose interaction failures missed by task outcomes.

\begin{table}[!t]
  \centering
  {\small
  \setlength{\tabcolsep}{1mm}
  \begin{tabular*}{\columnwidth}{@{\extracolsep{\fill}}lccc@{\hspace{1mm}}lccc@{}}
    \toprule
    \multicolumn{4}{c}{\textbf{(a) Human Agreement} \(\uparrow\)}
    & \multicolumn{4}{c}{\textbf{(b) IDM Micro-MSE} \(\downarrow\)} \\
    \cmidrule(r){1-4}\cmidrule(l){5-8}
    \textbf{Eval.} & \textbf{Exp.} & \textbf{OOD} & \textbf{Macro}
    & \textbf{Input} & \textbf{Exp.} & \textbf{Early} & \textbf{Cross} \\
    \midrule
    VLM & 0.883 & 0.450 & 0.666
    & Sim. ref. & 0.051 & 0.156 & 0.283 \\
    IDM & 0.759 & 0.284 & 0.522
    & Generated & 0.046 & 0.158 & 0.307 \\
    \textbf{RMFA} & \textbf{0.801} & \textbf{0.750} & \textbf{0.776}
    & & & & \\
    \bottomrule
  \end{tabular*}
  }
  \caption{Evaluator validity (\(N=750\)). \textbf{(a)} Human action-following
    agreement: binary for VLM and Spearman \(\rho\) for IDM/RMFA; Macro averages
    Expert and Diverse/OOD. \textbf{(b)} IDM Micro-MSE by control source under
    matched horizons.}
  \label{tab:evaluator_validity}
\end{table}

\subsection*{5.5 Benchmark Validity and Downstream Relevance (RQ4)}
\noindent\textbf{Agreement with human judgments.}
We compare RMFA with a binary VLM action-following judgment adapted from rollout
evaluation~\cite{li2025worldmodelbench,yang2026mirabench} and IDM recovery error,
following video-to-action evaluation~\cite{qin2025worldsimbench,jiang2026robowmbench}.
Human labels cover all 750 rollouts pooled across six ACWMs: 250 each from
expert, early-policy, and cross-action controls (success/failure: 250/0,
144/106, and 0/250; 394/356 overall). We report Expert separately from
Diverse/OOD (early-policy and cross-action). VLM agreement uses thresholded
ratings; IDM and RMFA use Spearman \(\rho\) against graded ratings.

The VLM predicts positive action following for \(91.7\%\) of Diverse/OOD samples versus \(42.2\%\) by humans, yielding \(0.450\) agreement. Under these controls, RMFA remains strongly aligned (\(\rho=0.750\)), whereas IDM falls to \(\rho=0.284\). On correct simulator references, IDM error is \(3.1\times\) and \(5.6\times\) higher for early-policy and cross-action controls, respectively. This dependence without ACWM error indicates inverse-model decoding failure rather than action-realization fidelity.

\begin{figure}[!h]  \centering  \includegraphics[width=0.94\columnwidth]{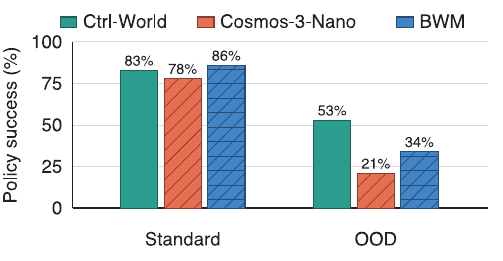}  \caption{Downstream utility of generated training data: policy  success is similar under standard controls but diverges under OOD  controls.}  \label{fig:downstream_synthetic_utility}
\end{figure}

\noindent\textbf{Synthetic-data utility.}
We conduct a controlled downstream study on one RoboTwin task. We collect expert task trajectories and construct simulator-validated counterfactual trajectories, then condition Ctrl-World, BWM, and Cosmos-3 on identical initial observations and action streams to generate matched synthetic training sets. Counterfactual rollout fidelity ranks Ctrl-World highest, followed by BWM and Cosmos-3. Policies trained on the corresponding synthetic data have standard trajectories clustered at \(78\)--\(86\%\) success, whereas OOD
trajectories separate Ctrl-World, BWM, and Cosmos-3-Nano at \(53\%\), \(34\%\), and
\(21\%\), respectively, consistent with the WorldSimProbe ordering
(Figure~\ref{fig:downstream_synthetic_utility}). Success-oriented training can
therefore obscure fidelity differences that matter beyond familiar controls.

\section*{6. Conclusion and Future Work}

We introduced WorldSimProbe, which reframes ACWM
evaluation from judging plausible or task-successful roll-
outs to testing capabilities required of physical simulators.
By standardizing evaluation around the chain from supplied
action to agent motion and grounded environment response,
five suites expose weaknesses overlooked by coarse, task-
associated protocols. Across six open-source ACWMs and
three platforms, the evaluation reveals that apparent action
responsiveness can mask miscalibrated realization, fidelity
degrades systematically under trajectory and behavior-source
shifts, and interaction failures exhibit recurring structure
across contact cues and dynamics primitives. Human judg-
ments and downstream policy performance support these
diagnoses. Beyond ranking models, WorldSimProbe reveals
where, when, and how simulator fidelity breaks down, pro-
viding a foundation for improving ACWMs as embodied
simulators. Currently focused on simulation, future work can
extend WorldSimProbe to real-world rollouts with physical
measurements, longer-horizon and closed-loop prediction,
and deformable or multi-object interactions; it may also guide
targeted data collection and model refinement.

\section*{Acknowledgments}
This work was supported by the National Natural Science Foundation of China (62476011), the Beijing Natural Science Foundation (L252060), and the Beijing Major Science and Technology Project under Contract no. Z191100010618003.

\FloatBarrier
\bibliography{aaai2027}

\appendix
\setcounter{figure}{0}
\setcounter{table}{0}
\setcounter{equation}{0}
\renewcommand{\thefigure}{S\arabic{figure}}
\renewcommand{\thetable}{S\arabic{table}}
\renewcommand{\theequation}{S\arabic{equation}}

\section*{Supplementary Material}
This supplement provides benchmark-construction details, evaluator implementations, experimental settings, and supporting analyses for WorldSimProbe. Sections~A--C document benchmark construction, evaluator implementation, and experimental setup. Section~D provides robustness analyses, Section~E documents human evaluation, and Section~F reports the downstream study. The main paper is self-contained; this document supplies supporting evidence and reproduction details.


\section{Benchmark Construction}
\label{sec:supp-construction}

\subsection{Platforms and Task Coverage}
WorldSimProbe spans three platforms selected for broad task and interaction diversity across single- and dual-arm embodiments. Table~\ref{tab:supp-platforms} summarizes their task coverage and reference-data configuration.

\begin{table*}[t]
\centering
\small
\setlength{\tabcolsep}{4pt}
\renewcommand{\arraystretch}{1.12}
\caption{Benchmark platforms and reference-data configurations. Only the external scene view is evaluated for consistent observability across ACWMs.}
\label{tab:supp-platforms}
\begin{tabular}{@{}p{0.10\textwidth}p{0.25\textwidth}p{0.14\textwidth}p{0.19\textwidth}p{0.24\textwidth}@{}}
\toprule
\textbf{Platform} & \textbf{Task coverage} & \textbf{Embodiment} & \textbf{Evaluated view} & \textbf{Action/video rate} \\
\midrule
\textbf{RoboTwin} & 50 bimanual manipulation tasks & Aloha-AgileX & \texttt{head\_camera}, $320\!\times\!240$ & Aligned actions/observations: $\sim$16.7~Hz; native video: 30~FPS \\
\addlinespace[2pt]
\textbf{ManiSkill} & 11 single-arm manipulation tasks & Panda & \texttt{base\_camera}, $128\!\times\!128$ & Actions/video: 20~Hz \\
\addlinespace[2pt]
\textbf{LIBERO} & 40 single-arm manipulation tasks & Panda & \texttt{agentview}, $128\!\times\!128$ & Actions/video: 20~Hz \\
\bottomrule
\end{tabular}
\end{table*}

\subsection{Task Data Generation}
The evaluated models are trained on the officially released platform training sets. All test trajectories are generated independently and are absent from the released training demonstrations. Because intervention semantics differ by task and suite, each task uses a dedicated generation script implemented through the corresponding simulator API. The shared pipeline samples a task and scene, generates a suite-specific intervention, executes it in simulation, applies reference-side validity checks, and records each accepted rollout and its metadata. The suite-specific procedures below detail the resulting units, interventions, coverage, and balancing policies.

\taskheading{T1: Local Action Calibration.}
\noindent\textbf{Construction.} Starting from a successful reference trajectory, we perturb one non-gripper action dimension over a fixed temporal window. For each triplet, we sample two distinct magnitudes from $\{0.0025, 0.005, 0.010, 0.015, 0.020\}$ in the simulator's native action units and assign the smaller and larger values to the small and large variants, respectively. The simulator seed, perturbation dimension, direction, and window are held fixed within each triplet.
\par\smallskip
\noindent\textbf{Validation and Selection.} Qualitative inspection confirmed that this range caused no readily discernible trajectory or outcome change while producing measurable differences in the simulator references. We retain only triplets whose three references remain task-valid and successful and satisfy $0 < D(x^0,x^s) < D(x^0,x^\ell)$. All filtering uses simulator references only.
\taskheading{T2: Global Trajectory Coverage.}
\noindent\textbf{Construction.} We sample a receiver scene and an action trajectory from a different donor task, then execute the complete donor stream from the receiver's initial state.
\par\smallskip
\noindent\textbf{Validation and Selection.} Simulator replay supplies the counterfactual reference. A receiver--donor pair enters the manifest only when the full replay is executable and passes the common reference-validity checks.

\taskheading{T3: Action-Source Behavior Preservation.}
\noindent\textbf{Construction.} For each of five selected tasks per platform, we form matched groups sharing the task, episode, and scene seed. The original successful reference trajectory constitutes the expert source.
\par
\begin{figure*}[t]
  \centering
  \includegraphics[width=0.98\textwidth]{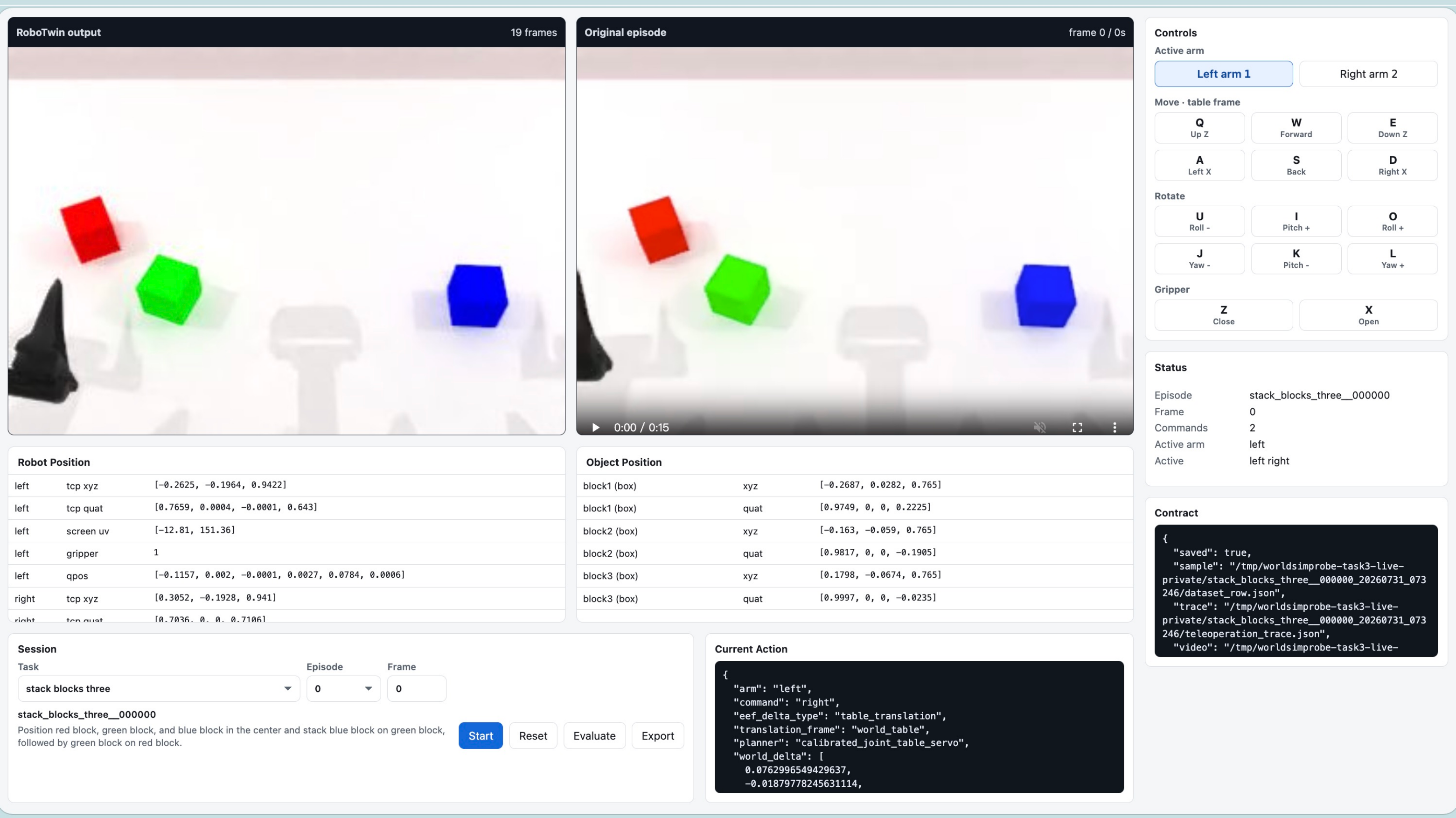}
  \caption{Online teleoperation interface used to collect Task 3 human trajectories. Operators view the episode's expert rollout and use buttons mapped to robot controls to reproduce the demonstrated behavior.}
  \label{fig:supp-teleoperation-interface}
\end{figure*}

\smallskip
\noindent\textbf{Human Teleoperation.} Five operators view the expert rollout video for each episode and independently imitate it using an online teleoperation interface whose buttons map to robot controls (Figure~\ref{fig:supp-teleoperation-interface}). Operators are not stratified by experience, and collection quality is not separately scored; the resulting trajectories preserve naturally occurring differences among independent executions of the same demonstrated behavior.
\par\smallskip
\noindent\textbf{Policy Sources.} We train $\pi_{0.5}$~\cite{intelligence2025pi0_5} on the training split of the same five tasks and use checkpoints at 5k and 50k training steps as the early- and late-policy sources, respectively.
\par\smallskip
\noindent\textbf{Validation and Selection.} Each human, policy, and expert trajectory that passes the common reference-validity checks is retained as one instance. Matching the underlying episode and scene isolates source-specific behavior from scene variation.

\taskheading{T4: Interaction Grounding.}
\noindent\textbf{Design Motivation.} In a qualitative audit of 50 observed interaction-grounding failures across the evaluated ACWMs, all 50 (100\%) were false positives, in which the rollout generated an unsupported interaction. We observed no false negatives, in which supported contact failed to elicit an environment response. Within this error sample, ACWMs therefore showed a strong tendency to hallucinate rather than omit interactions. Task~4 consequently isolates false-positive grounding through three controlled no-contact interventions.
\par\smallskip
\noindent\textbf{Distractor object.} We insert a plausible object from a same-task donor episode into the receiver scene while replaying the original action stream. This tests whether scene context overrides the supplied controls, causing the model to redirect agent motion toward the distractor, or whether the distractor responds despite interaction remaining grounded in the original target.\par\smallskip\noindent\textbf{Appearance-induced false contact.} We replay the original arm trajectory while clamping both grippers open. A model that relies on task or visual priors rather than the supplied controls may hallucinate gripper closure and the resulting object interaction.\par\smallskip\noindent\textbf{Spatial proximity.} We relocate the dominant movable target beyond the robot's control-induced swept region while preserving the original action stream. Failure occurs if the model redirects agent motion toward the relocated object or generates an environment response unsupported by realized contact.
\par\smallskip
\noindent\textbf{Validation and Selection.} Distractor cases are retained only when the inserted object is stable, in workspace, sufficiently separated from existing geometry, and neither contacted nor displaced while the receiver task remains successful. Open-gripper cases require the commanded grippers to remain open and the dominant target to remain stable. Target-shift cases require a stable, collision-free placement and no robot contact. All selection decisions use simulator state before model inference.

\taskheading{T5: Interaction Dynamics.}
\noindent\textbf{Construction.} Each instance begins from a successful simulator trajectory. We identify the longest usable gripper--object contact segment, select a compatible task--object pair, branch near initial contact, and replace the remaining controls with a scripted primitive: planar side-contact push; articulated pull; grasped planar drag; tangential in-place rotation; three-cycle grasped shake; brief press-and-retract tap; high lateral knock-over; or grasp--lift--release drop.
\par\smallskip
\noindent\textbf{Validation and Selection.} Simulator-state criteria verify that the intended primitive occurred. The defaults require \(xy\) displacement of at least \(0.015\,\mathrm{m}\) for push; yaw change of at least \(0.10\,\mathrm{rad}\) for rotate; lift and fall of at least \(0.04\,\mathrm{m}\) each for drop; task success or articulation change of at least \(0.05\) for pull and tap; horizontal displacement of at least \(0.03\,\mathrm{m}\) or yaw range of at least \(0.08\,\mathrm{rad}\), with \(z\)-range at most \(0.04\,\mathrm{m}\), for shake; \(xy\) displacement of at least \(0.025\,\mathrm{m}\), with \(z\)-range at most \(0.04\,\mathrm{m}\), for drag; and final tilt of at least \(0.50\,\mathrm{rad}\) with tilt increase of at least \(0.35\,\mathrm{rad}\) for knock-over. Accepted instances are balanced by primitive within each platform: 233 per primitive over all eight primitives in RoboTwin, 295 per primitive over all eight in ManiSkill, and 500 per primitive over push, pull, tap, and drop in LIBERO.

Rollout horizons are not globally fixed: each instance retains its task- and episode-specific action duration, and generated rollouts are aligned to the corresponding reference horizon and frame rate.

\subsection{Reference Validation and Filtering}
Candidate instances are executed completely in the simulator before inclusion. An instance is retained only when the controller accepts the full action stream, joint and workspace limits are satisfied, no numerical instability occurs, no unintended collision occurs outside the prescribed interaction, and the suite-specific success, contact, or no-contact condition is satisfied. Simulator crashes, incomplete rollouts, missing observations, invalid controls, failed tracking, and references that do not realize the required condition are removed. The manifest counts below are measured after all common and suite-specific checks and are fixed before model inference.

\subsection{Observations, Actions, and Alignment}
As summarized in Table~\ref{tab:supp-platforms}, evaluation uses only the external scene view for fairness because most evaluated ACWMs support a single external or head camera. Reference videos remain at native resolution and are resized only during model- or evaluator-specific preprocessing.

Each instance stores synchronized joint-space and end-effector action streams. A model receives the representation used during its pretraining, defaulting to joint actions otherwise. Preliminary interface checks across models found no consistent advantage for joint- or end-effector-space conditioning; these checks informed the native-interface choice but are not included in benchmark scoring. For every instance, all six ACWMs receive the same initial observation, action stream, simulator seed, reference horizon, and three shared diffusion seeds. Reported scores first average the three stochastic generations before higher-level aggregation. Evaluator-specific temporal synchronization and rate normalization are detailed in Section~\ref{sec:supp-temporal}.

\subsection{Final Test Manifest}
Table~\ref{tab:supp-inventory} reports the final 18,608 controlled evaluation instances. These counts describe only the filtered simulator-derived test set and exclude model outputs and stochastic repetitions.

\begin{table*}[t]
\centering
\small
\renewcommand{\arraystretch}{1.02}
\caption{Final filtered test-manifest composition.}
\label{tab:supp-inventory}
\begin{tabular*}{\textwidth}{@{\extracolsep{\fill}}lrrrrrr}
\toprule
\textbf{Platform} & \textbf{Suite 1} & \textbf{Suite 2} & \textbf{Suite 3} & \textbf{Suite 4} & \textbf{Suite 5} & \textbf{Total} \\
\midrule
\textbf{RoboTwin} & 419 & 1,575 & 1,000 & 640 & 1,864 & 5,498 \\
\textbf{ManiSkill} & 1,000 & 1,500 & 750 & 1,000 & 2,360 & 6,610 \\
\textbf{LIBERO} & 1,000 & 1,500 & 1,000 & 1,000 & 2,000 & 6,500 \\
\midrule
\textbf{Total} & \textbf{2,419} & \textbf{4,575} & \textbf{2,750} & \textbf{2,640} & \textbf{6,224} & \textbf{18,608} \\
\bottomrule
\end{tabular*}
\end{table*}

The manifest records the platform, task, suite, scene seed, embodiment, synchronized joint-space and end-effector action streams, intervention parameters, reference video, robot and object states, and contact state. We will release the complete filtered test set, generation and teleoperation scripts, manifests, action streams, simulator-reference videos and metadata, and evaluator implementations. Generated ACWM outputs and checkpoints are not redistributed because the evaluated models are independently available as open-source releases; their outputs can be reproduced using the provided test data, manifests, and evaluation recipes.

\section{Evaluator Implementation}
\label{sec:supp-evaluators}

\subsection{Temporal Synchronization and Rate Normalization}\label{sec:supp-temporal}

All simulator-reference and generated rollouts share the benchmark-defined start time $t=0$. For video $X$, let $X_{i,j}$ denote frame $j$ of instance $i$, with timestamp $\tau^X_{i,j}$. Explicit per-frame timestamps are used when available; otherwise, timestamps are inferred from the native frame rate $f_X$ as $\tau^X_{i,j}=j/f_X$. Temporal normalization uses only these timestamps; we do not apply dynamic time warping, content-based phase shifting, or learned frame interpolation. For target timestamp $t_k$, nearest-frame sampling is\begin{equation}j_X(k)=\arg\min_j |\tau^X_{i,j}-t_k|,\qquad\widetilde X_i(t_k)=X_{i,j_X(k)}.\end{equation}

This changes only the sampling rate and does not reorder frames or compensate for model-response delays.

\subsection{Task 1 Temporal Alignment and MSE}
For Task~1, the generated original rollout $\hat{x}_i^0$ defines the comparison timebase. For perturbation variant $v\in\{s,\ell\}$, we retain original-rollout timestamps within the pair's common duration:\begin{equation}\mathcal T_i^{0,v}=\{\tau^0_{i,k}:\tau^0_{i,k}\leq\min(T_i^0,T_i^v)\}.\end{equation}

At each $t_k\in\mathcal T_i^{0,v}$, the nearest frame from the perturbation rollout is selected. Temporally aligned full-video MSE is\begin{equation}\begin{aligned}D(\hat{x}_i^0,\hat{x}_i^v)&=\frac{1}{|\mathcal T_i^{0,v}|HWC}\sum_{t_k\in\mathcal T_i^{0,v}}\sum_{p,c}\\[-2pt]&\quad \times\left(\hat{x}_i^0(t_k,p,c)-\widetilde{\hat{x}}_i^v(t_k,p,c)\right)^2.\end{aligned}\end{equation}

The original--small and original--large distances enter the Task~1 calibration score. If the generated original--large distance is zero, the response ratio is undefined and the instance receives score zero; otherwise, the ratio and oracle-relative score follow the main-paper definition. Because all three rollouts share $t=0$ and the intended prediction horizon, the comparison measures perturbation responses at matched physical times despite native-rate differences.

\begin{figure}[t]
    \centering
    \includegraphics[width=\columnwidth]{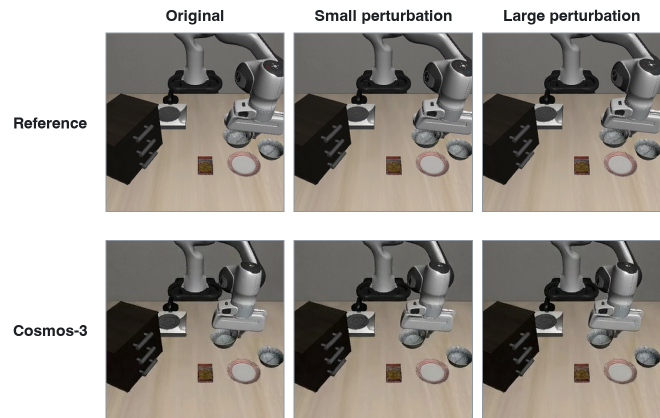}
    \caption{Representative Task~1 local-calibration triplet. Simulator-reference and Cosmos-3 frames are shown at a matched time for the original, small-, and large-perturbation variants.}
    \label{fig:supp-task1-demo}
\end{figure}

\subsection{Robot-Masked Flow Alignment}
For Tasks~2 and~3, simulator reference $x_i^*$ and generated rollout $\hat{x}_i$ are sampled on a fixed evaluation timebase $f_e=5\,\mathrm{Hz}$, with $\Delta t=0.2\,\mathrm{s}$. Over their common duration $T_i=\min(T_i^*,\hat{T}_i)$, the evaluation grid and aligned frames are\begin{equation}\begin{aligned}\mathcal T_i&=\{t_k=k/f_e:0\leq t_k\leq T_i\},\\[-2pt]\bar{x}_{i,k}^*&=\widetilde{x}_i^*(t_k),\qquad\bar{\hat{x}}_{i,k}=\widetilde{\hat{x}}_i(t_k).\end{aligned}\end{equation}

Reference-arm masks are generated by RobotSeg~\cite{mei2026robotseg} using the \texttt{robotseg.pt} checkpoint, while dense optical flow is estimated by DPFlow~\cite{morimitsu2025dpflow} using the \texttt{things} checkpoint. Both use pretrained checkpoints without task-specific fine-tuning; masks and flows are evaluated on the shared $160\times120$ grid, and RobotSeg is applied only to simulator-reference frames.

Dense optical flow between consecutive aligned frames is\begin{equation}F_{i,k}^*=\operatorname{Flow}(\bar{x}_{i,k}^*,\bar{x}_{i,k+1}^*),\qquad\hat{F}_{i,k}=\operatorname{Flow}(\bar{\hat{x}}_{i,k},\bar{\hat{x}}_{i,k+1}),\end{equation}so both flow sequences represent motion over the same nominal $200$-ms interval.

RMFA uses overlapping $2$-s windows with a $1$-s stride. At $5$ Hz, each window contains ten consecutive flow fields. Let $\mathcal K_{i,w}$ denote the flow indices in window $w$. The reference-arm mask for each flow pair is the union of its endpoint masks:
\begin{equation}A_{i,k}^*=A(\bar{x}_{i,k}^*)\lor A(\bar{x}_{i,k+1}^*).\end{equation}

The active spatiotemporal robot-motion mask is
\begin{equation}\mathcal M_{i,w}=\left\{(k,p):\begin{aligned}k&\in\mathcal K_{i,w},\ A_{i,k}^*(p)=1,\\[-2pt]&\|F_{i,k}^*(p)\|_2>\tau\end{aligned}\right\}.\end{equation}

Within this mask, RMFA computes\begin{align}E_{i,w}&=\left[\begin{aligned}&\frac{1}{|\mathcal M_{i,w}|}\sum_{(k,p)\in\mathcal M_{i,w}}\\[-2pt]&\quad \|\hat{F}_{i,k}(p)-F_{i,k}^*(p)\|_2^2\end{aligned}\right]^{1/2},\\R_{i,w}&=\left[\begin{aligned}&\frac{1}{|\mathcal M_{i,w}|}\sum_{(k,p)\in\mathcal M_{i,w}}\\[-2pt]&\quad \|F_{i,k}^*(p)\|_2^2\end{aligned}\right]^{1/2},\\S_{i,w}&=100\max\left(0,1-\frac{E_{i,w}}{\max(R_{i,w},c)}\right).\end{align}

At the shared $160\times120$ flow resolution, we use $\tau=0.25$ pixels and $c=3.16$ pixels. These reference-calibrated constants are fixed across platforms and evaluated models. Task~2 first averages $S_{i,w}$ across windows within each instance and then across instances; Task~3 applies the same procedure within each action-source group.

\begin{figure*}[t]
    \centering
    \includegraphics[width=0.90\textwidth]{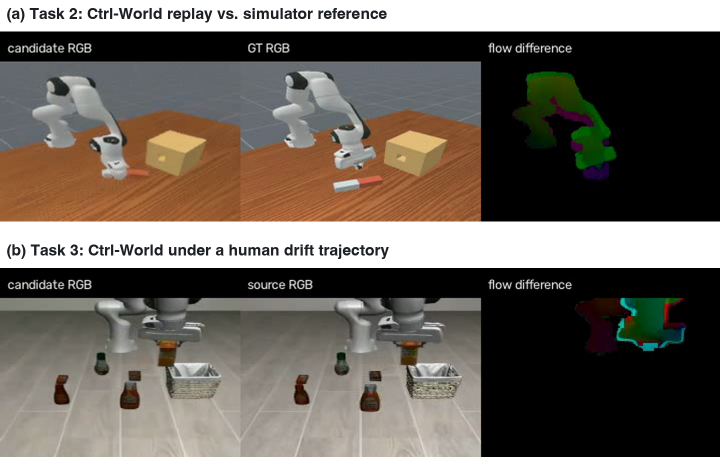}
    \caption{Representative RMFA evaluator views. (a) Task~2 counterfactual replay and its simulator reference. (b) Task~3 human-source trajectory and its generated rollout. Each row shows the candidate frame, matched reference or source frame, and robot-region flow difference.}
    \label{fig:supp-task23-demo}
\end{figure*}

\subsection{Interaction-Grounding Tracking}
TAPNext++~\cite{jung2026tapnextpp} uses the \texttt{tapnextpp\_ckpt.pt} checkpoint and receives frames resized to $256\times256$ and normalized to $[-1,1]$. Evaluated object poses (and distractor poses when present) are projected with the stored camera extrinsic and intrinsic matrices, and a $3\times3$ query grid with a 10-pixel radius is initialized around each projected center. Query points with visibility probability below $0.5$ are ignored; if no point is visible in a frame, the previous centroid is carried forward. The minimum visible-frame fraction is recorded as tracker reliability but does not affect the score.

\textbf{Scoring.} For the robot-motion gate, $\hat g_i$ is the configurable maximum robot-arm centroid displacement across the three gate frames. A generated rollout passes when $\hat g_i$ reaches minimum displacement in the canonical $256\times256$ coordinate system. Among gate-passing rollouts, displacement of the evaluated object by more than 10 pixels constitutes an unsupported interaction. Both thresholds are fixed from simulator-reference diagnostics. Motion-gate or tracking failure receives zero:
\begin{equation}
S_i=100\,\mathbf{1}[\hat g_i\ge60]\,\mathbf{1}[d_i\le10].
\end{equation}

\begin{figure}[t]
    \centering
    \includegraphics[width=\columnwidth]{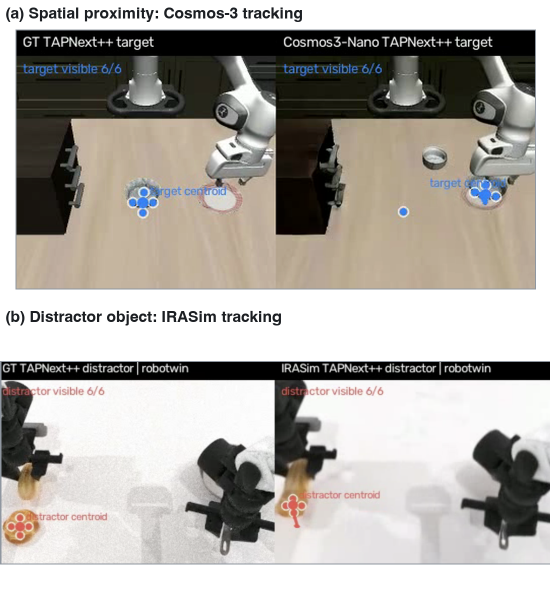}
    \caption{Representative Task~4 TAPNext++ tracking views for spatial-proximity and distractor-object interventions. Query points and tracked centroids are overlaid on simulator-reference and generated frames.}
    \label{fig:supp-task4-demo}
\end{figure}

\subsection{Interaction-Primitive Evaluator}
We use Qwen3-VL-8B-Instruct~\cite{bai2025qwen3} as a deterministic judge. Each generated rollout is uniformly sampled into 12 chronological frames spanning the full candidate clip and supplied as ordered images. Inference uses bfloat16 precision, \texttt{do\_sample=False}, and \texttt{max\_new\_tokens=384}. The prompt defines the expected agent motion and object response for each primitive and requires a structured JSON response containing one primitive label and separate binary agent- and object-motion judgments. Neither the intended primitive nor the ACWM identity is provided to the judge.

\begin{figure*}[p]
    \centering
    \includegraphics[width=0.90\textwidth]{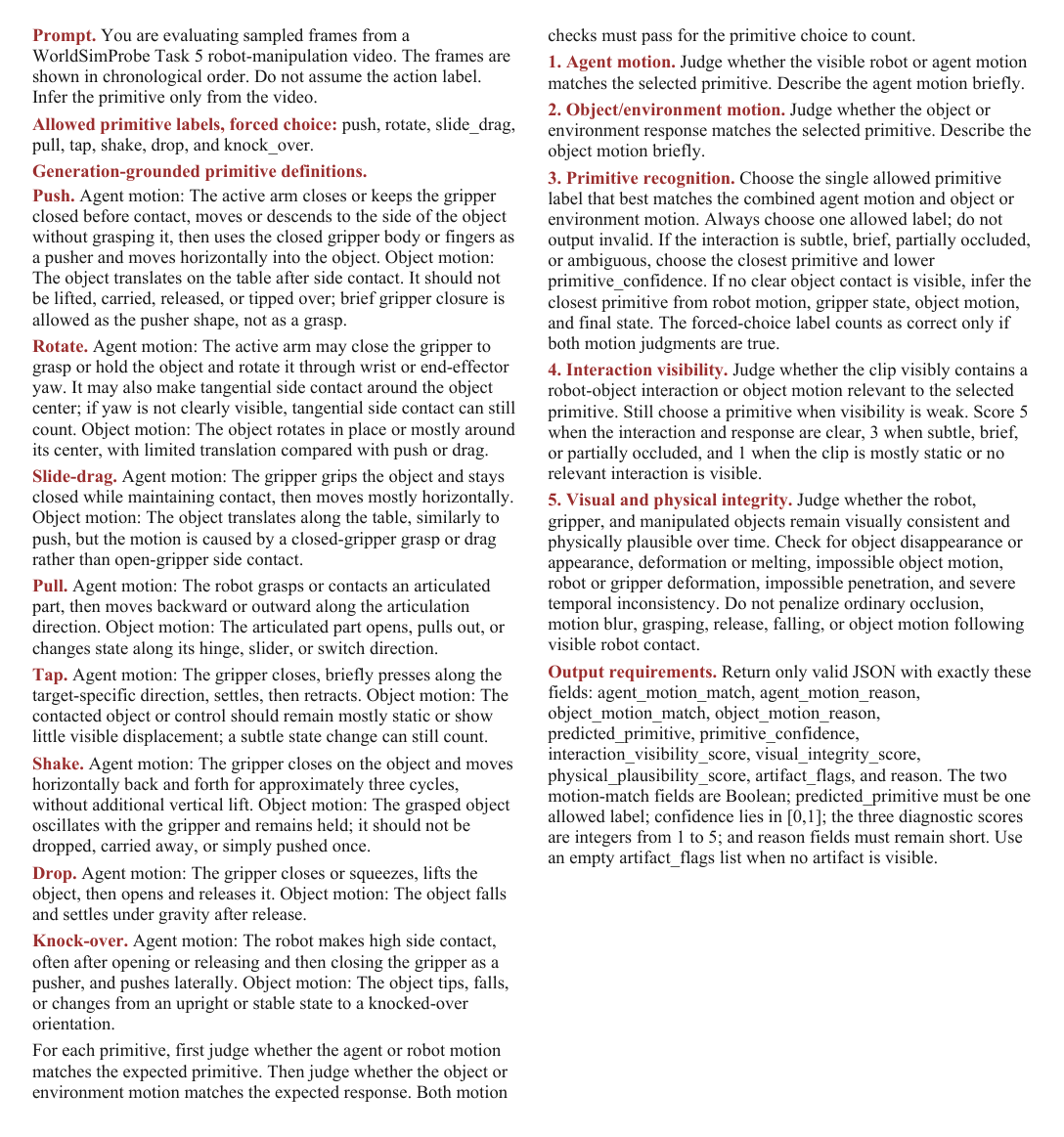}
    \caption{Task 5 VLM evaluator prompt.}
    \label{fig:task5-vlm-prompt}
\end{figure*}

Let $p_i$ and $\hat p_i$ denote the intended and predicted primitives, and let $a_i,o_i\in\{0,1\}$ indicate whether the agent- and object-motion judgments satisfy the primitive specification. The implementation-level Task~5 score is
\begin{equation}
S_i=100\,\mathbf{1}[\hat p_i=p_i]\,\mathbf{1}[a_i=1]\,\mathbf{1}[o_i=1].
\end{equation}
Responses from which a valid primitive label cannot be parsed receive zero. Scores are averaged across instances within each primitive and then across primitives. The main text's $\hat{\pi}_i$ denotes the accepted primitive prediction: the label is accepted only when both motion judgments pass; otherwise, the instance is counted as incorrect. Because instances are balanced across primitives within each platform, the main text's instance average is equivalent to the primitive-macro average used here. To validate simulator-defined primitive labels, we human-annotate a stratified sample of 1,864 simulator-reference Task~5 rollouts: 560 from RoboTwin, 704 from ManiSkill, and 600 from LIBERO. This sample represents 29.95\% of the full reference pool. Each rollout was independently labeled by three annotators. For all 1,864 rollouts, at least two annotators assigned the same primitive label, yielding 100\% consensus coverage. The complete judge prompt and response schema appear in Figure~\ref{fig:task5-vlm-prompt}.

\begin{figure}[t]
    \centering
    \includegraphics[width=\columnwidth]{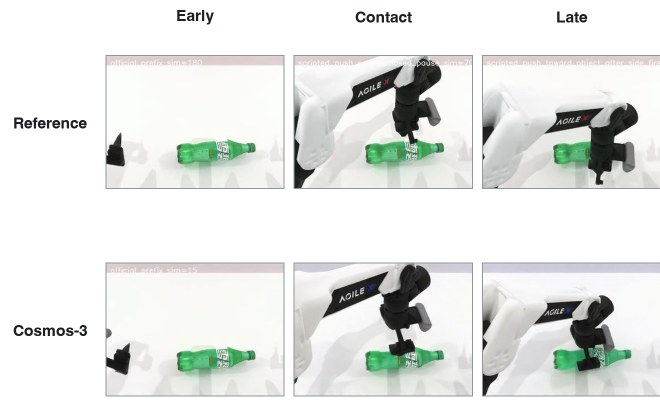}
    \caption{Representative Task~5 push rollout. Simulator-reference and Cosmos-3 frames are shown before contact, at contact, and late in the interaction.}
    \label{fig:supp-task5-demo}
\end{figure}

\subsection{VLM and IDM Baselines}

\textbf{VLM.}
We use the same Qwen3-VL-8B-Instruct checkpoint and 12 uniformly sampled frames as in Task~5. The model judges whether each rollout follows the supplied action specification. Human ratings range from 1 to 5 and are binarized as positive at ratings $\geq 3$ for agreement analysis.

\textbf{IDM.}
We use the official CLAM implementation of a VPT-style inverse dynamics model~\cite{baker2022vpt}. Its CNN encoder maps adjacent $84\times84$ RGB frames $(x_t,x_{t+1})$ to action $a_t$, using context length 1 and a 128-D embedding. We train one model per platform on the same official training split used for the ACWMs, with native outputs of 14-D joint control for RoboTwin, 8-D joint control for ManiSkill, and 7-D end-effector control for LIBERO, including grippers. All three platform-specific IDMs use a global batch size of 1,024 and 25,000 training steps. Reference and generated clips use matched timestamps and horizons. Micro-MSE averages squared action error over matched timesteps and dimensions, and human agreement is measured by Spearman's $\rho$ between negative Micro-MSE and the mean human rating.

\begin{figure*}[!t]
\centering
\small
\setlength{\tabcolsep}{3.4pt}
\renewcommand{\arraystretch}{1.00}
\captionof{table}{ACWM configurations: (a) released training and inference recipes; (b) platform-specific action interfaces and source-frame geometry.}
\label{tab:supp-models}
\textbf{(a) Resolved released recipes}\par
\begin{tabular}{@{}l p{0.19\textwidth} p{0.225\textwidth} c c p{0.17\textwidth}@{}}
\toprule
\textbf{Model} & \textbf{Initialization; trainable modules} & \textbf{Optimizer; LR schedule} & \textbf{Batch/acc.} & \textbf{Precision} & \textbf{Inference} \\
\midrule
IRASim & IRASim-XL/2; transformer (VAE frozen) & AdamW, $10^{-4}$; constant & $1/1$ & FP32 & PNDM, 50 steps, $g=1$ \\
Ctrl-World & SVD; U-Net and action encoder & AdamW, $10^{-5}$; constant & $4/1$ & FP16 & 50 steps, $g=1$ \\
BWM & Wan2.2-TI2V-5B; DiT and action encoder & AdamW, $5\!\times\!10^{-5}$; constant & $1/8$ & BF16 & 50 steps, $g=1$ \\
DreamDojo & DreamDojo-2B; DiT and action embedder & AdamW, $10^{-4}$; 1k warmup, then constant & $8/1$ & BF16 & 35 steps, $g=0$ \\
LingBot-VA & Released checkpoint; full action--video transformer & AdamW, $10^{-5}$; warmup, then constant & $1/1$ & BF16 & 25 video steps; video CFG $=5$ \\
Cosmos-3 & Cosmos-3-Nano; released action-modality modules & FusedAdam, $2\!\times\!10^{-4}$; LambdaLinear & $1/1$; pack $\leq32$ & BF16 & 30 steps, $g=1$ \\
\bottomrule
\end{tabular}

\textbf{(b) Platform-specific interfaces}\par
\begin{tabular}{@{}p{0.13\textwidth} p{0.16\textwidth} p{0.40\textwidth} c c@{}}
\toprule
\textbf{Platform} & \textbf{Embodiment} & \textbf{Native action supplied to ACWM} & \textbf{Dim.} & \textbf{Source frames} \\
\midrule
RoboTwin & Aloha-AgileX & Absolute joint positions and two grippers & 14 & $320\!\times\!240$ \\
ManiSkill & Panda & Seven joint positions and gripper & 8 & $128^2$ \\
LIBERO & Panda & End-effector delta pose and gripper & 7 & $128^2$ \\
\bottomrule
\end{tabular}

\end{figure*}

\section{Experimental Setup}
\label{sec:supp-setup}

\subsection{Model Training and Inference}
We train one checkpoint per model--platform pair on the official training split using each baseline's released software environment and training scripts. Only platform interfaces are adapted; LingBot-VA and Cosmos-3 remain action-conditioned but use video-only decoding. Table~\ref{tab:supp-models} summarizes the resolved recipes and adaptations.

All models receive the same initial observation, native action trajectory, simulator seed, prediction horizon, and three shared generation seeds. We average the three rollout scores per instance before higher-level aggregation.

\subsection{Compute and Software}
Experiments were conducted on three identically configured nodes, each containing eight NVIDIA H100 80GB GPUs, two Intel Xeon Platinum 8462Y+ CPUs, and 2~TiB RAM, running Ubuntu 22.04.5.

\section{Robustness and Additional Analysis}
\label{sec:supp-robustness}

\subsection{Visual-Quality Robustness Across Tasks}
\label{sec:supp-visual-quality}

Visual quality is not an explicit benchmark objective. As a robustness diagnostic, we report the change in MUSIQ relative to matched simulator references, defined as generated-video MUSIQ minus reference-video MUSIQ; more negative values indicate greater degradation.

\begin{table}[t]
\centering
\small
\setlength{\tabcolsep}{3.5pt}
\caption{MUSIQ change relative to matched simulator references across tasks. More negative values indicate greater visual-quality degradation.}
\label{tab:supp-musiq-by-task}
\begin{tabular}{@{}clcc@{}}
\toprule
\textbf{Task} & \textbf{Evaluator} & \textbf{$\Delta$MUSIQ} & \textbf{95\% CI} \\
\midrule
T1 & MSE      & $-9.74$  & $[-10.36,-9.13]$ \\
T2 & RMFA     & $-10.97$ & $[-11.80,-10.18]$ \\
T3 & RMFA     & $-10.16$ & $[-11.04,-9.24]$ \\
T4 & Tracking & $-9.57$  & $[-10.44,-8.69]$ \\
T5 & VLM      & $-8.90$  & $[-9.45,-8.34]$ \\
\bottomrule
\end{tabular}
\end{table}

Task~2 exhibits the greatest multi-model degradation, motivating a targeted test of whether RMFA remains sensitive to the affected visual evidence.

\subsection{RMFA Sensitivity in Task 2}
\label{sec:supp-rmfa-visual-sensitivity}

\begin{figure*}[!t]
    \centering
    \includegraphics[width=\textwidth]{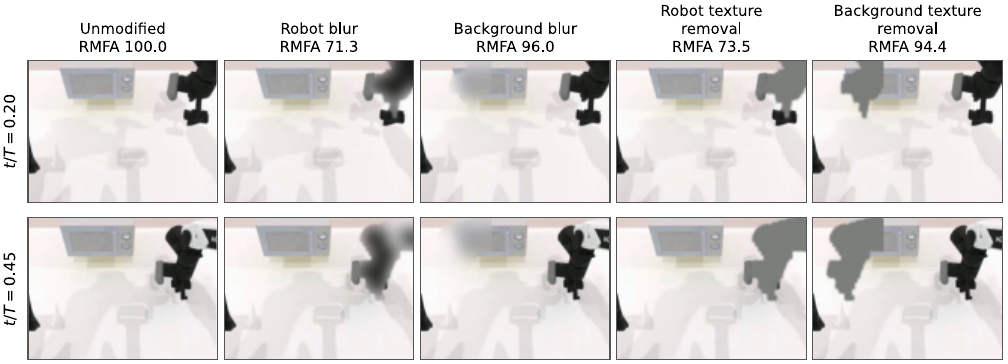}
    \caption{Representative Task~2 masked-corruption example. The same simulator reference video and timestamps are shown across all conditions. Gaussian blur ($\sigma=12$) and complete texture removal ($\alpha=1$) are applied either within the robot mask or to an equal-area background region; the RMFA score for each condition is shown above the corresponding frames.}
    \label{fig:supp-rmfa-corruption-example}
\end{figure*}

Following Table~\ref{tab:supp-musiq-by-task}, we test whether RMFA responds to visual degradation that obscures task-relevant robot motion. For each of 50 simulator Task~2 reference videos, we apply Gaussian blur and texture removal at matched strengths either inside the robot segmentation mask or to an equal-area region sampled outside that mask. The strongest settings, Gaussian blur with $\sigma=12$ and complete texture removal with $\alpha=1$, are shown in Figure~\ref{fig:supp-rmfa-corruption-example}. Action trajectories, timestamps, and all uncorrupted pixels remain identical within each pair. We report the RMFA decrease from the unmodified reference; 95\% paired bootstrap confidence intervals are obtained by resampling the 50 reference trajectories with replacement.

\begin{table}[t]
\centering
\small
\setlength{\tabcolsep}{4pt}
\caption{RMFA response to controlled visual corruption in Task~2.}
\label{tab:supp-rmfa-corruption}
\begin{tabular}{@{}lcc@{}}
\toprule
\textbf{Condition} & \textbf{RMFA drop} & \textbf{95\% CI} \\
\midrule
Robot-region corruption & $30.55$ & $[28.35,32.83]$ \\
Background control & $7.93$ & $[6.99,8.90]$ \\
Paired excess drop & $\mathbf{22.62}$ & $\mathbf{[20.98,24.35]}$ \\
\bottomrule
\end{tabular}
\end{table}

RMFA decreases monotonically as robot-region corruption strength increases in 48/50 trajectories. The equal-area background control estimates RMFA's response to generic image degradation; the paired difference isolates the additional effect of corrupting robot-motion evidence.

\section{Human Evaluation}
\label{sec:supp-human}
Human evaluation covers all 750 rollouts pooled across the six ACWMs. The set contains 250 expert, 250 early-policy, and 250 cross-action rollouts, with the success/failure composition shown in Table~\ref{tab:supp-human-composition}. Every rollout receives a graded action-following label.

\textbf{Annotation protocol.} Three annotators independently rated every rollout. For each item, the matched generated and simulator-reference videos were presented together, and annotators rated whether the generated video reproduced the reference behavior on a 1--5 scale. Scores 1--2 indicate failure to follow the reference robot action or motion; a score of 3 indicates that robot action following is preserved even if the environment response differs; scores 4--5 indicate increasingly close agreement in both robot motion and environment response. Thus, faithful action following is a prerequisite for a rating of at least 3. For each rollout, we average the three ratings to obtain one graded human score; this mean is used directly for rank correlation and thresholded at $\geq3$ for binary agreement. Model identity and control source were hidden, and presentation order was randomized.

\begin{table}[!ht]
\centering
\small
\renewcommand{\arraystretch}{1.08}
\caption{Composition of the 750-rollout human-evaluation set.}
\label{tab:supp-human-composition}
\begin{tabular}{@{}lrrr@{}}
\toprule
\textbf{Control source} & \textbf{Total} & \textbf{Success} & \textbf{Failure} \\
\midrule
Expert & 250 & 250 & 0 \\
Early policy & 250 & 144 & 106 \\
Cross action & 250 & 0 & 250 \\
\midrule
Total & 750 & 394 & 356 \\
\bottomrule
\end{tabular}
\end{table}

\section{Downstream Synthetic-Data Study}
\label{sec:supp-downstream}

We extend the three-model downstream study reported in the main paper to all six evaluated ACWMs on the same RoboTwin task. Expert task trajectories and simulator-validated counterfactual trajectories are used to construct matched synthetic training sets. Table~\ref{tab:supp-downstream} reports policy success under standard and OOD controls for this expanded comparison.

\begin{table}[!h]
\centering
\small
\renewcommand{\arraystretch}{1.08}
\caption{Downstream policy success (\%) using synthetic training data from each ACWM. Higher is better; best results are bold.}
\label{tab:supp-downstream}
\begin{tabular}{@{}lcc@{}}
\toprule
\textbf{ACWM} & \textbf{Standard $\uparrow$} & \textbf{OOD $\uparrow$} \\
\midrule
IRASim      & 86 & 40 \\
Ctrl-World  & 83 & \textbf{53} \\
BWM         & 86 & 34 \\
DreamDojo   & 81 & 22 \\
LingBot-VA  & \textbf{89} & \textbf{53} \\
Cosmos-3    & 78 & 21 \\
\bottomrule
\end{tabular}
\end{table}

Success is relatively compressed under standard controls ($78$--$89\%$) but diverges under OOD controls ($21$--$53\%$). We interpret this experiment as a controlled case study of the practical relevance of the diagnosed fidelity differences rather than universal downstream validation.


\FloatBarrier

\begin{figure*}[p]
    \centering
    \includegraphics[page=1,width=0.92\textwidth]{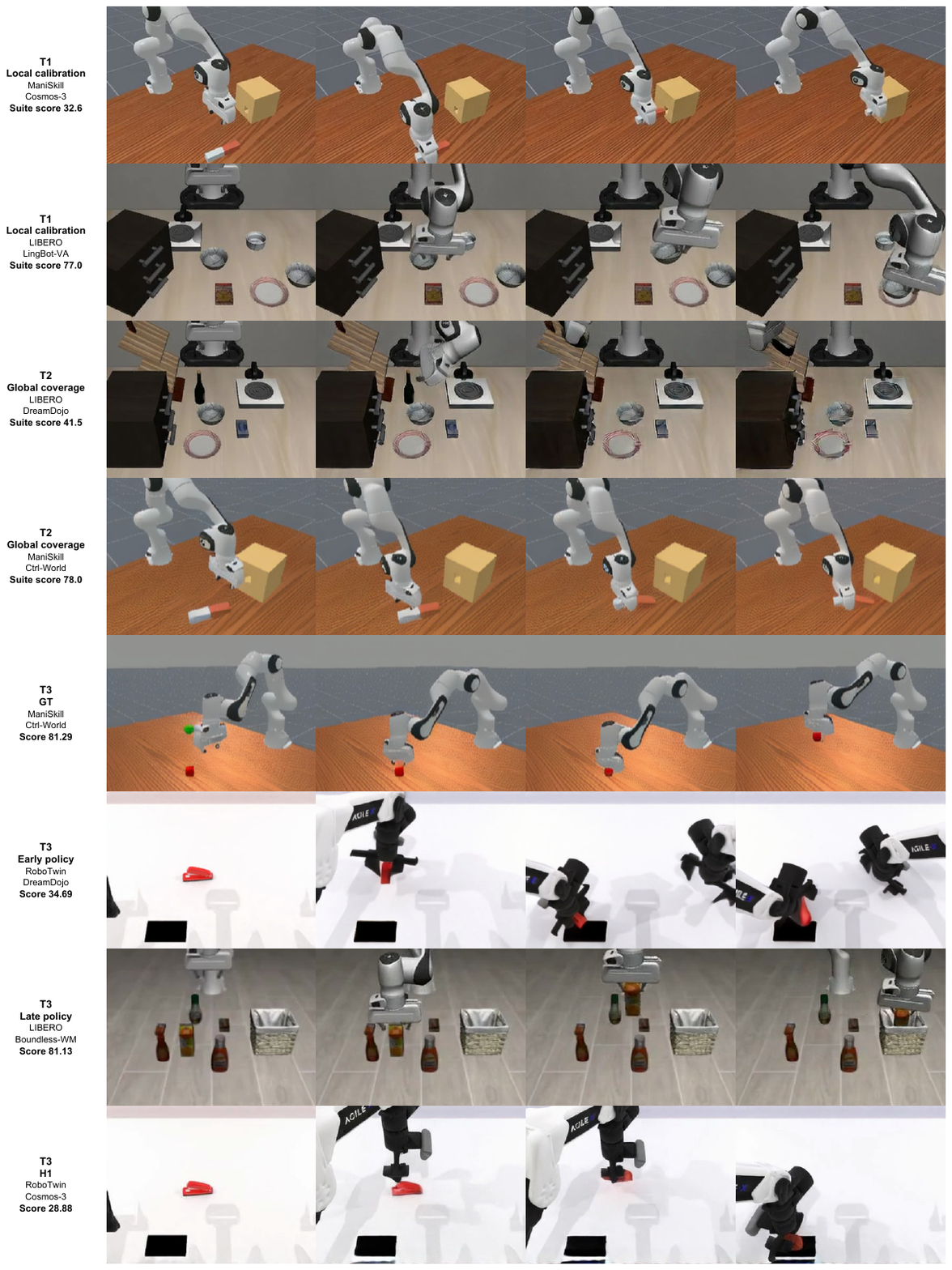}
    \caption{Representative scored qualitative rollouts for Tasks 1--3. Each row shows four frames sampled from the beginning to the end of one rollout. Tasks 1--4 report the corresponding evaluator score; human teleoperators are anonymized as H1--H5.}
    \label{fig:scored-qualitative-tasks123}
\end{figure*}

\begin{figure*}[p]
    \centering
    \includegraphics[page=2,width=0.92\textwidth]{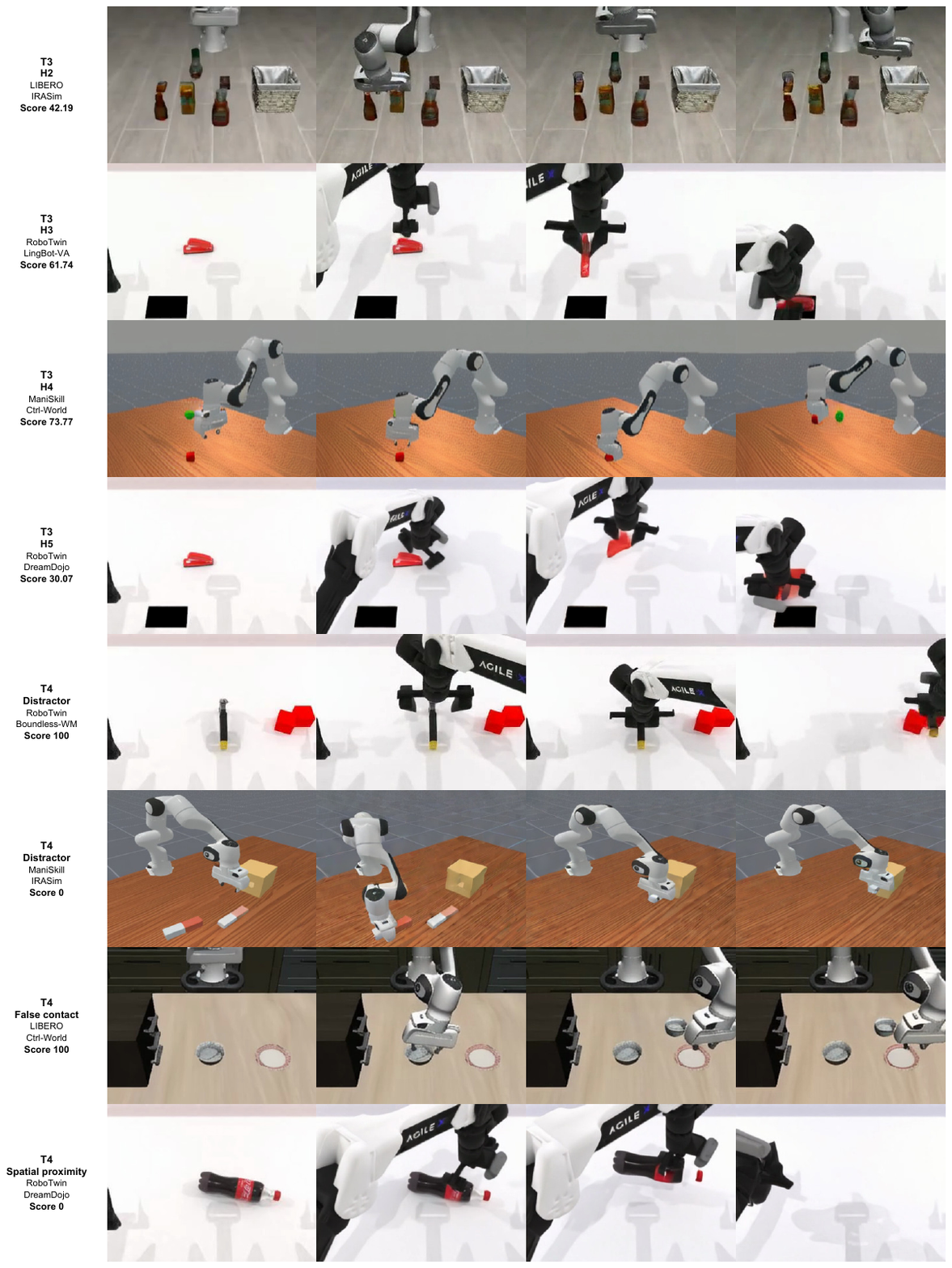}
    \caption{Representative scored qualitative rollouts for source-diverse actions and interaction grounding. Task 4 rows cover distractor-object, appearance-induced false-contact, and spatial-proximity interventions.}
    \label{fig:scored-qualitative-tasks34}
\end{figure*}

\begin{figure*}[p]
    \centering
    \includegraphics[page=3,width=0.92\textwidth]{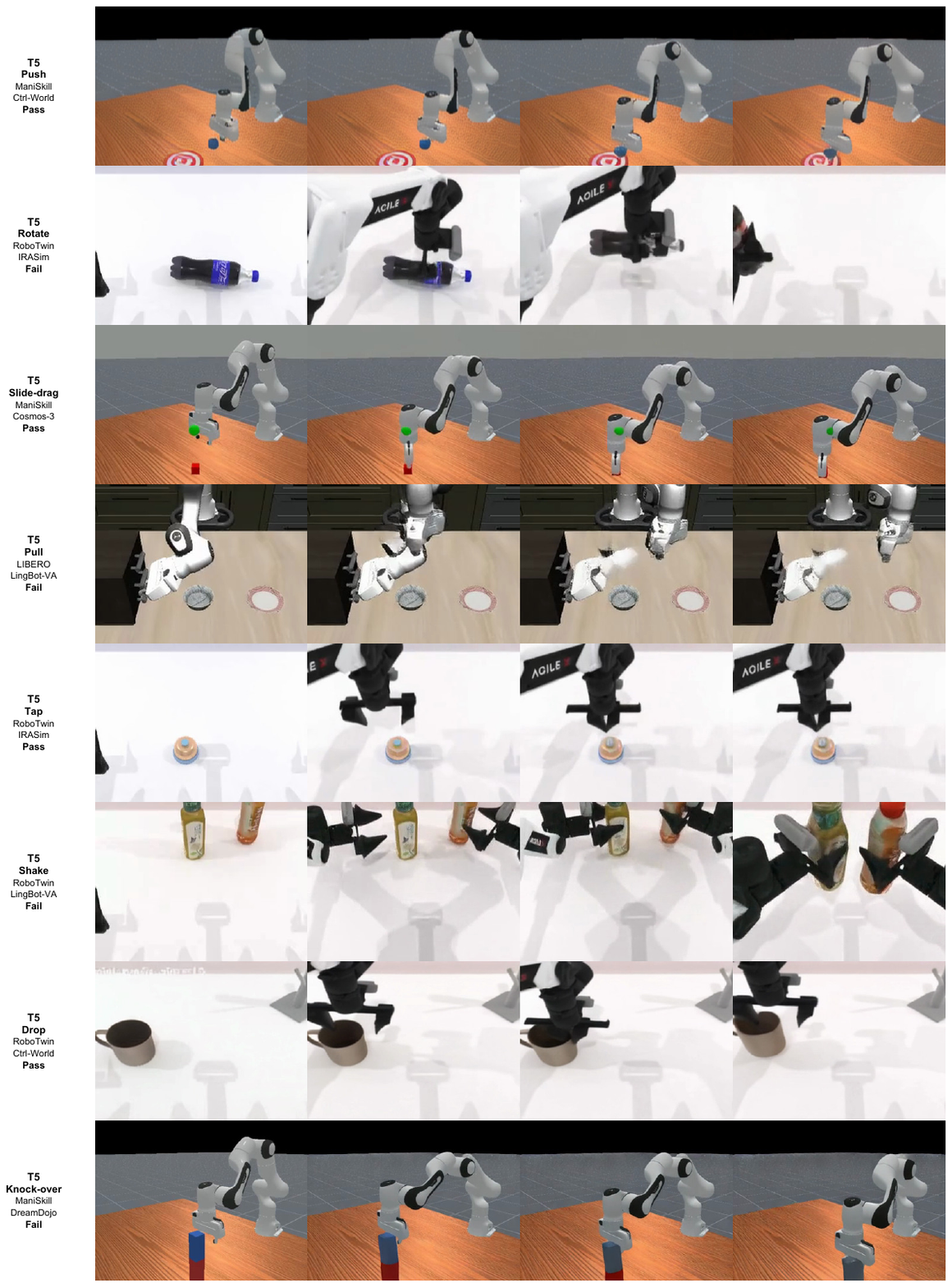}
    \caption{Representative Task 5 interaction-dynamics rollouts spanning the evaluated primitives. Each row shows four frames sampled from the beginning to the end of one rollout together with the evaluator pass/fail decision.}
    \label{fig:scored-qualitative-task5}
\end{figure*}

\FloatBarrier

\end{document}